\documentclass[journal]{IEEEtran}
\usepackage{physics}
\usepackage{graphicx}
\usepackage{float}
\usepackage{amsmath}
\usepackage{subfigure}
\usepackage{amsmath}
\usepackage{longtable}
\usepackage{booktabs}
\usepackage{float}
\usepackage{threeparttable}
\usepackage{xcolor}
\usepackage{diagbox}
\usepackage{cite}
\usepackage{hyperref}
\usepackage{color}
\usepackage{amsfonts}
\usepackage{amsmath}
\usepackage{amssymb}
\usepackage{booktabs}
\usepackage{multirow}

 \renewcommand{\paragraph}[1]{
    \vspace{2mm}
     \noindent\textbf{#1} 
 }

 \usepackage{bm}
\ifCLASSINFOpdf
\else
\fi

\begin{document}

\title{Domain Adaptation based Object Detection for Autonomous Driving in Foggy and Rainy Weather}

\author{Jinlong Li, Runsheng Xu, Xinyu Liu, Jin Ma, Baolu Li, Qin Zou, Jiaqi Ma, Hongkai Yu$^{*}$ 

\thanks{Jinlong Li, Jin Ma, and Xinyu Liu are with the Department of  Computer Science, Cleveland State University, Cleveland, OH 44115, USA. Baolu Li and Hongkai Yu are with the Department of Electrical and Computer Engineering, Cleveland State University, Cleveland, OH 44115, USA. Runsheng Xu and Jiaqi Ma are with the Department of Civil and Environmental Engineering, University of California, Los Angeles, CA 90024, USA. Qin Zou is with the School of Computer Science, Wuhan University, Wuhan 430072, China. A preliminary version of this work has been published on the IEEE/CVF WACV2023 conference~\cite{li2023domain_wacv}. This work was supported by NSF 2215388.} 
\thanks{* Corresponding author: Hongkai Yu (e-mail: h.yu19@csuohio.edu).} 
}

\maketitle 

 \begin{abstract}
Typically, object detection methods for autonomous driving that rely on supervised learning make the assumption of a consistent feature distribution between the training and testing data, this such assumption may fail in different weather conditions. Due to the domain gap, a detection model trained under clear weather may not perform well in foggy and rainy conditions. Overcoming detection bottlenecks in foggy and rainy weather is a real challenge for autonomous vehicles deployed in the wild. To bridge the domain gap and improve the performance of object detection in foggy and rainy weather, this paper presents a novel framework for domain-adaptive object detection. The adaptations at both the image-level and object-level are intended to minimize the differences in image style and object appearance between domains. Furthermore, in order to improve the model's performance on challenging examples, we introduce a novel adversarial gradient reversal layer that conducts adversarial mining on difficult instances in addition to domain adaptation. Additionally, we suggest generating an auxiliary domain through data augmentation to enforce a new domain-level metric regularization. Experimental findings on public benchmark exhibit a substantial enhancement in object detection specifically for foggy and rainy driving scenarios.
The code is available at \url{https://github.com/jinlong17/DA-Detect}.
\end{abstract}

\begin{IEEEkeywords}
intelligent vehicles, deep learning, object detection, domain adaptation
\end{IEEEkeywords}

\IEEEpeerreviewmaketitle

\section{Introduction}

\IEEEPARstart{T}{he} past decade has witnessed the significant breakthroughs on autonomous driving with artificial intelligence methods~\cite{9963987,chen2022parallel}, leading to numerous applications in transportation, including improving traffic safety~\cite{nie2020multimodality,parseh2021data,10036095}, reducing traffic  congestion~\cite{10079130,gao2021novel}, minimizing air pollution~\cite{9508812,hong2021co}, and enhancing traffic efficiency~\cite{10026339,9930673,10100881}. Object detection is a critical component of autonomous driving, which relies on computer vision and artificial intelligence techniques to understand driving scenarios~\cite{10093116,9963987}. However, the foggy and rainy weather conditions make the understanding of camera images particularly difficult, which poses challenges to the camera based object detection system installed on the intelligent  vehicles~\cite{9815132,10045043,xu2021opencda}.

\begin{figure}[!t]
\centering
\subfigure[The domain gap between different weather in the wild]{%
  \includegraphics[width=1\columnwidth]{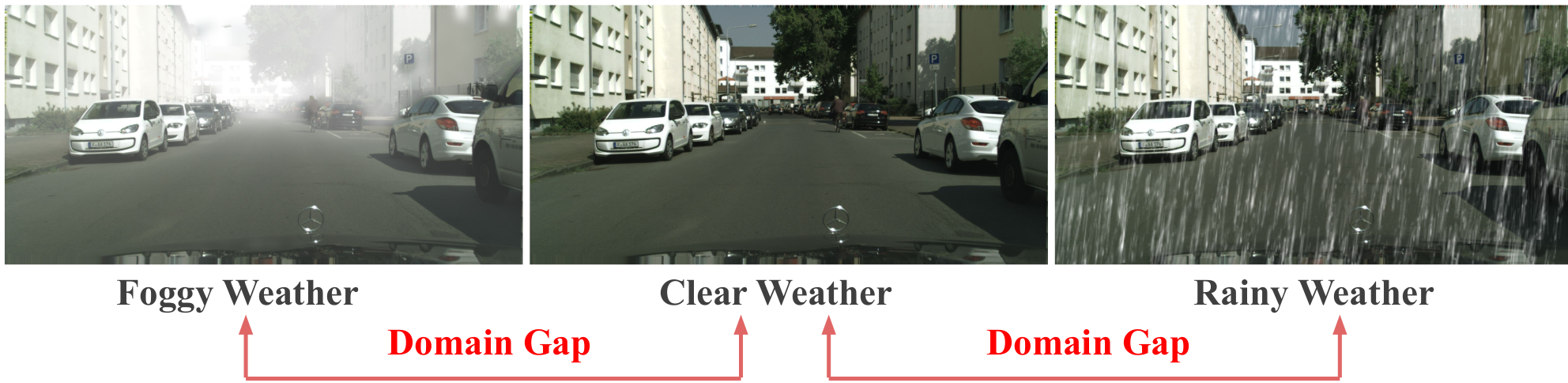}%
}
\hfil
\subfigure[Detection in clear weather]{%
  \includegraphics[width=0.48\columnwidth]{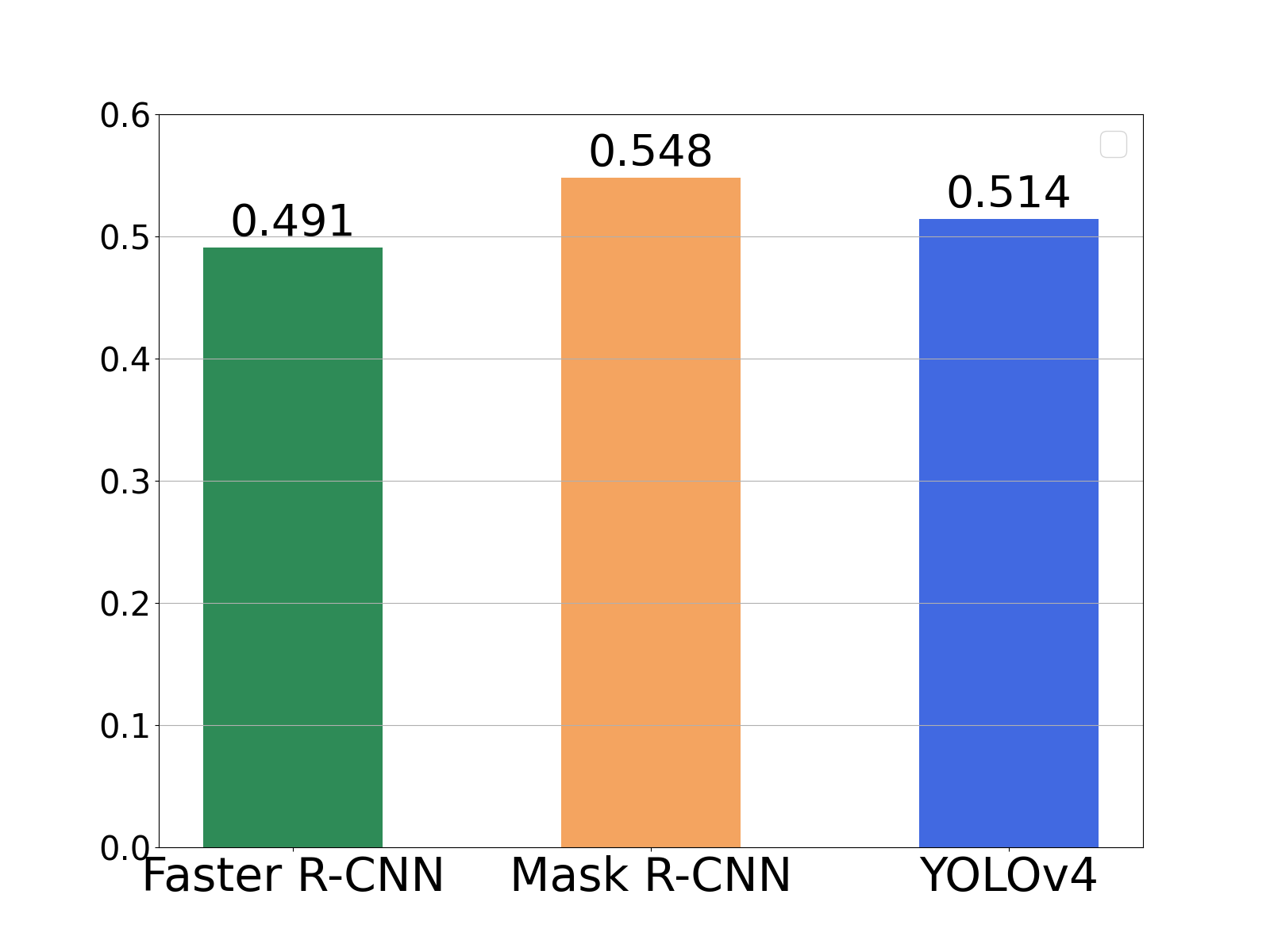}%
}
\hfil
\subfigure[Detection in foggy weather]{%
  \includegraphics[width=0.48\columnwidth]{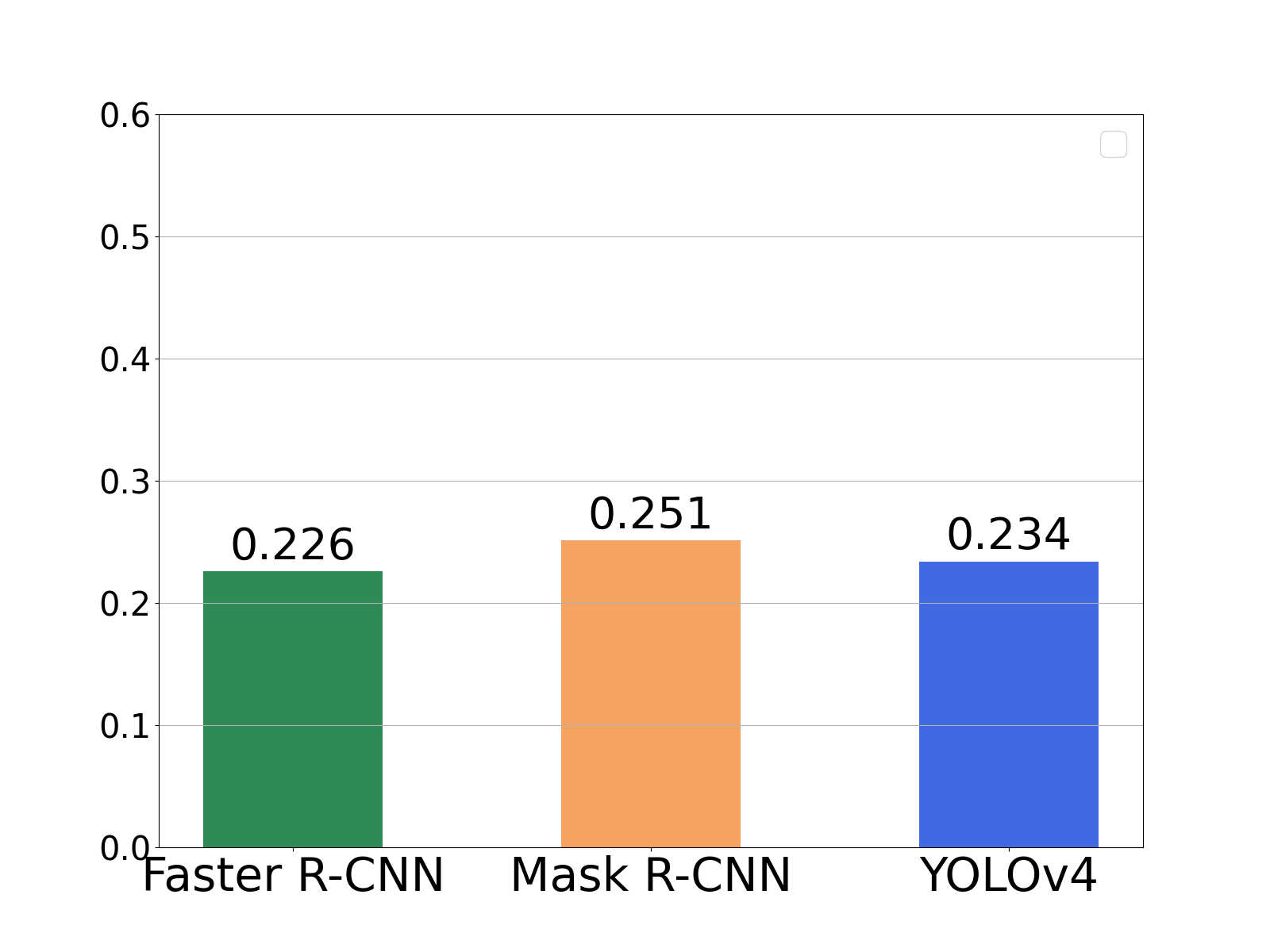}%
}
\caption{Illustration of the weather domain gap (foggy and rainy) for autonomous driving and the detection performance drop because of the domain gap. Three deep learning  models ( Faster R-CNN~\cite{ren2015faster}, Mask R-CNN~\cite{he2017mask}, and YOLOv4~\cite{bochkovskiy2020yolov4}) are all trained with the clear weather data of   Cityscapes~\cite{cordts2016cityscapes}.}
\label{fig:motivation}
\end{figure}

Thanks to the rapid advancements in deep learning, numerous object detection deep learning-based methods have achieved remarkable success in intelligent transportation systems. However, the impressive performance of these popular methods heavily relies on large-scale annotated data for supervised learning. Moreover, these methods make the assumption of consistent feature distributions between the training and testing data. In reality, this assumption may not hold true, especially in diverse weather conditions~\cite{sakaridis2018semantic}. For example, as depicted in Fig.~\ref{fig:motivation}, CNN models such as YOLOv4~\cite{bochkovskiy2020yolov4}, Faster R-CNN~\cite{ren2015faster}, and Mask R-CNN~\cite{he2017mask}, trained on clear-weather data (source domain), exhibit accurate object detection performance under clear weather conditions (Fig.~\ref{fig:motivation}b). However, their performance significantly degrades under foggy weather conditions (Fig.~\ref{fig:motivation}c). This degradation can be attributed to the presence of a feature domain gap between different weather conditions, as illustrated in Fig.~\ref{fig:motivation}a. The model trained on the source domain is not familiar with the feature distribution in the target domain. Consequently, this paper aims to enhance object detection specifically in foggy and rainy weather conditions through domain adaptation-based transfer learning.

The objective of this paper is to reduce the domain gap between various weathers for enhanced object detection. To handle the domain shift problem (\textit{e.g.} Clear$\to$Foggy and  Clear$\to$Rainy), in this paper, we present a new domain adaptation framework that aims to enhance the robustness of object detection in foggy and rainy weather conditions. Our proposed framework follows an unsupervised setting, similar to previous works~\cite{chen2018domain,song2020multi,li2021domain}. In this setting, we have well-labeled clear-weather images as the source domain, while the foggy and rainy weather images, which serve as the target domains, lack any annotations.
This unsupervised setting is because adverse weather images with labeling (manual annotating) are time-consuming and costly. Inspired by~\cite{chen2018domain,9439889},  
the proposed method aims to reduce the domain feature discrepancies in both image style and object appearance.
To enhance robustness and prevent data-level overfitting, we propose a Dynamic Masking Process to generate masked images by ``dropping" some pixel regions.
To achieve domain-invariant features, 
we incorporate both image-level and object-level domain classifiers as components to facilitate domain adaptation in our CNN architecture. 
These classifiers are responsible for distinguishing between different domains. By employing an adversarial approach, our detection model learns to generate features that are invariant to domain variations, thereby confusing the domain classifiers. This adversarial design encourages the network to produce features that are agnostic to specific weather conditions, leading to improved object detection performance in foggy and rainy weather scenarios.

Furthermore, we propose a novel methodology for domain adaptation (DA). Current existing domain adaptation methods~\cite{chen2018domain,li2021domain,9435348,zheng2020cross,9439889} might ignore: 1) the different challenging levels of various training samples, 2) the domain-level feature metric distance to the third related domain by only involving the source domain and target domain. 
This paper investigates the incorporation of hard example mining and an additional related domain to 
further strengthen the model's ability to learn robust and transferable representations. 
We propose a novel Adversarial Gradient Reversal Layer (AdvGRL) and introduce an auxiliary domain through data augmentation. The AdvGRL is designed to perform adversarial mining on challenging examples, thereby improving the model's ability to learn in challenging scenarios. Additionally, the auxiliary domain is leveraged to enforce a new domain-level metric regularization during the transfer learning process.
In summary, the contributions of this paper can be summarized as follows:
\begin{itemize}
    \item This paper proposes a novel unsupervised domain adaptation method to enhance object detection for autonomous vehicles under foggy and rainy conditions, including the image-level and object-level adaptations.   
    
    \item This paper proposes to perform adversarial mining for hard examples during domain adaptation to further improve the model's transfer learning capabilities under challenging samples, which is accomplished by our proposed AdvGRL.  
    
    \item This paper proposes a new domain-level metric regularization to improve transfer learning, \textit{i.e.}, the regularization constraint between source domain, added auxiliary domain, and target domain.  

    \item This paper explores the intensive transfer learning experiments of clear$\rightarrow$foggy, clear$\rightarrow$rainy, cross-camera adaptation, and also carefully studies the different-intensity (small, medium, large) fog and rain adaptations.
    
\end{itemize}

\section{Related Work}\label{Sec:Related_Work}

\subsection{Detection for intelligent vehicles}
The contemporary realm of intelligent vehicles has garnered considerable attention, primarily directing towards the enhancement of road safety, mitigation of traffic congestion, and the overall optimization of transportation systems~\cite{9847095,10068744}.
Recent strides in deep learning have been pivotal in propelling the field of intelligent vehicles forward~\cite{li2020sus,xu2021holistic,10077757}. Within this landscape, object detection has emerged as a focal point of extensive research endeavors, encompassing the identification and classification of objects such as vehicles, pedestrians, traffic signs, traffic lights, and assessing road conditions~\cite{shan2019pixel,zhao2020fusion}.
Deep learning methods have been prominently introduced to address object detection tasks, generally falling into two distinct categories: two-stage object detectors and one-stage object detectors.
Faster RCNN~\cite{ren2015faster} and Mask RCNN~\cite{he2017mask} are one of the classic two-stage methods, which typically consist of two main stages: region proposal generation and object classification/localization.
While YOLO series~\cite{redmon2016you} and SSD~\cite{liu2016ssd} are one of the representative one-stage methods,  which typically use a set of predefined anchor boxes or default boxes at different scales and aspect ratios to densely cover the image.
\cite{10007064} designed an edge intelligence-based vehicle detection algorithm based on YOLOv4 to augment vehicle detection capabilities. \cite{hoffmann2020real}, on the other hand, proposed a multistage algorithm that initially leverages the YOLOv3 network for object detection. It's also worth noting that Faster R-CNN and Mask R-CNN have been conventionally employed for vehicle detection in the context of intelligent vehicles~\cite{chen2021deep}, attesting to their commendable performance in various scenarios. However, the direct application of these methods in autonomous driving settings is often constrained by the formidable challenges posed by adverse real-world weather conditions.

\subsection{Detection for intelligent vehicles under foggy and rainy weather} 
In recent years, considerable research has been dedicated to addressing the challenges posed by various weather conditions encountered in autonomous driving scenarios. Researchers have generated various datasets~\cite{sakaridis2018semantic,9247499} and proposed numerous methods~\cite{huang2020dsnet,hahner2021fog,qian2021robust,bijelic2020seeing,sindagi2020prior,9669124} to improve object detection under adverse weather conditions. One notable example is the Foggy Cityscape dataset, which is a synthetic dataset created by applying fog simulation to the Cityscape dataset~\cite{sakaridis2018semantic}.
In the context of object detection research in rainy weather, several synthesized rainy datasets have been proposed~\cite{hnewa2020object, sindagi2020prior, 9669124}. 
\cite{hahner2021fog}  devised a fog simulation technique to augment existing real lidar datasets, thereby enhancing their quality and realism.
The simulated foggy data offers valuable opportunities to enhance object detection methods that are specifically tailored for foggy weather conditions.
For leveraging information from multiple sensors,
\cite{bijelic2020seeing} designed a network to integrate data from different sensors \textit{e.g.}, LiDAR, camera, and radar. 
\cite{qian2021robust} proposed a method that exploits both LiDAR and radar signals to obtain object proposals. 
The features extracted from the regions of interest in both sensors are fused together to improve the performance of object detection.
However, these mentioned methods often rely on input data from different types of sensors other than the camera alone, which may not be applicable to all autonomous driving vehicles. Therefore, the objective of this work is to develop a DA network by utilizing only camera-sensor data as input.

\subsection{Object Detection via Domain Adaptation}
Domain adaptation is effective in reducing the distribution discrepancy between different domains, enabling models trained on a labeled source domain to be applicable to an unlabeled target domain.
There has been a growing interest in addressing domain adaptation for object detection~\cite{chen2018domain,kim2019diversify,saito2019strong,zhao2020adaptive,xu2021spg,zhang2021rpn,xu2020cross,9439889} in recent years. 
Several studies~\cite{chen2018domain,he2019multi,saito2019strong,xu2020cross,9439889} have explored the alignment of features from different domains to achieve DA object detectors. 
A DA Faster R-CNN framework~\cite{chen2018domain}  was proposed to reduce the domain gap at both the image level and instance level. 
He et al.~\cite{he2019multi} proposed a multi-adversarial network that aligns domain features and proposal features hierarchically to minimize domain distribution disparity. 
In addition to feature alignment,
image style transfer approaches~\cite{shan2019pixel,kim2019diversify,hsu2020progressive,schutera2020night} are utilized to address the challenge of DA.
An image translation module~\cite{shan2019pixel} was utilized to convert images from the source domain to the target domain. They then trained the object detector using adversarial training on the target domain. \cite{hsu2020progressive} adopted a progressive image translation strategy and introduced a weighted task loss during adversarial training to address image quality differences.
Several previous methods~\cite{zhou2022multi, li2022sigma,chen2022learning,wang2022parallel}  have also proposed complex architectures for domain adaptation in object detection. 
Feature Pyramid Networks (FPN) was utilized to incorporate pixel-level and category-level adaptation for object detection~\cite{zhou2022multi}. 
In order to incorporate the uncertainty of unlabeled target data,  \cite{chen2022learning} introduced an uncertainty-guided self-training mechanism, which leverages a Probabilistic Teacher and Focal Loss.
Different with these methods, our approach does not introduce additional learnable parameters to the Faster R-CNN. Instead, we utilize an AdvGRL and a Domain-level Metric Regularization based on triplet loss.
A key difference between our method and previous domain adaptation approaches lies in the treatment of training samples. 
While existing methods often assume that training samples are at the same challenging level, our approach introduces the AdvGRL for adversarial hard example mining,
specifically targeting the improvement of transfer learning performance. Additionally, to mitigate overfitting and improve domain adaptation, an auxiliary domain is generated and incorporated into domain-level metric regularization.

\section{Methodology}\label{Sec:Method}

This section introduces the overall network architecture, each detailed component, loss functions of our proposed method.

\subsection{Network Architecture}

\begin{figure*}[htb]
	\begin{minipage}[b]{1\textwidth}
		\centering
		\includegraphics[width=1\textwidth]{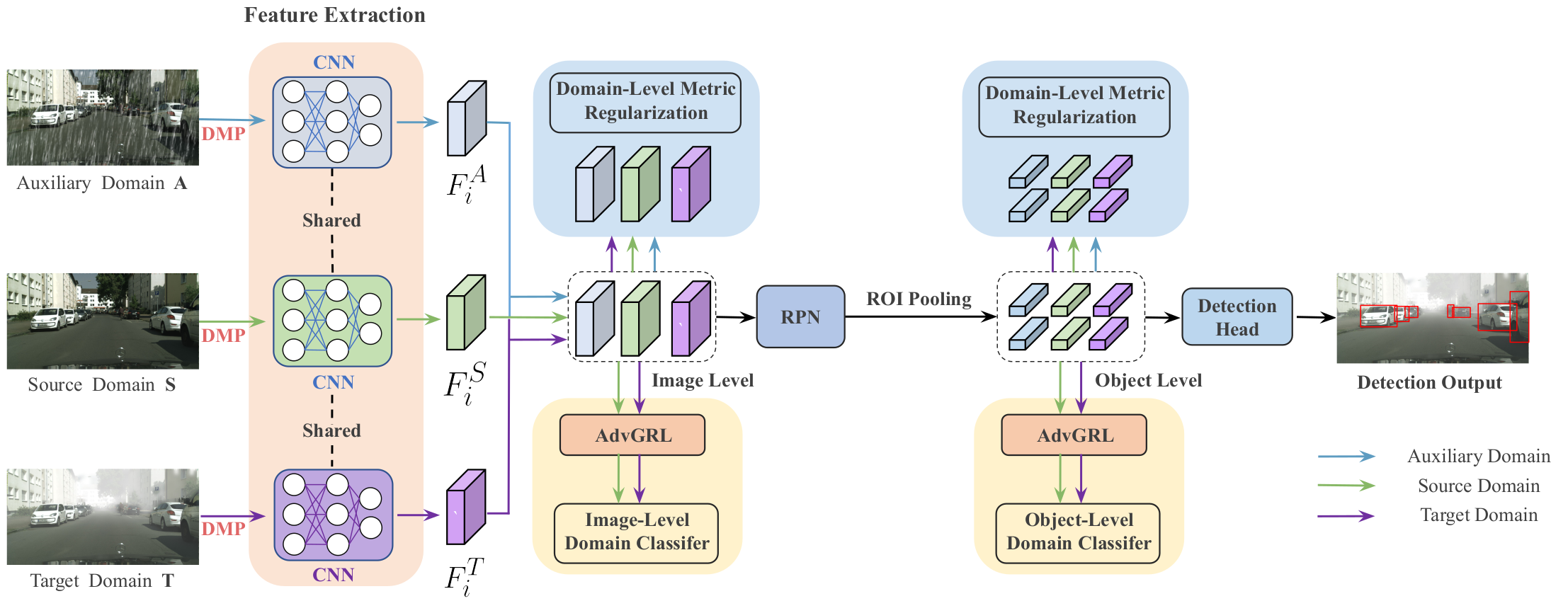}
	\end{minipage}
    \caption{The architecture of the proposed domain adaptation-based enhanced detection for intelligent vehicles in  foggy and rainy weather. Here we illustrate our target domain using the example of foggy weather. It is recommended to view this figure in color.}
	\label{fig:da_faster_rcnn} 
\end{figure*}

As shown in Fig.~\ref{fig:da_faster_rcnn}, our proposed network follows the pipeline of Faster R-CNN. In the first step, we deploy a Dynamic Masking Process to generate the masked images, then
we involve a CNN backbone to extract the image-level features from masked images.
These features are then fed into the Region Proposal Network to produce region proposals. 
The next stage involves the Region of Interest (ROI) pooling, both the image-level features and the object proposals are as input to obtain object-level features. 
Finally, we apply a detection head for the object-level features to make the final outputs. To enhance the framework of Faster R-CNN for domain adaptation, we incorporate two additional domain adaptation modules: image-level and object-level modules. Both of them utilize a novel AdvGRL in conjunction with the domain classifier. 
By combining these modules, we are able to extract domain-invariant features and effectively perform adversarial hard example mining.
Additionally, an auxiliary domain is introduced to enforce a new domain-level metric regularization.
During training, source, target, and auxiliary domains, are simultaneously utilized.

\subsection{Dynamic Masking Process}
\begin{figure}[!t]
\centering
\subfigure[]{%
  \includegraphics[width=0.32\columnwidth, height=0.6in]{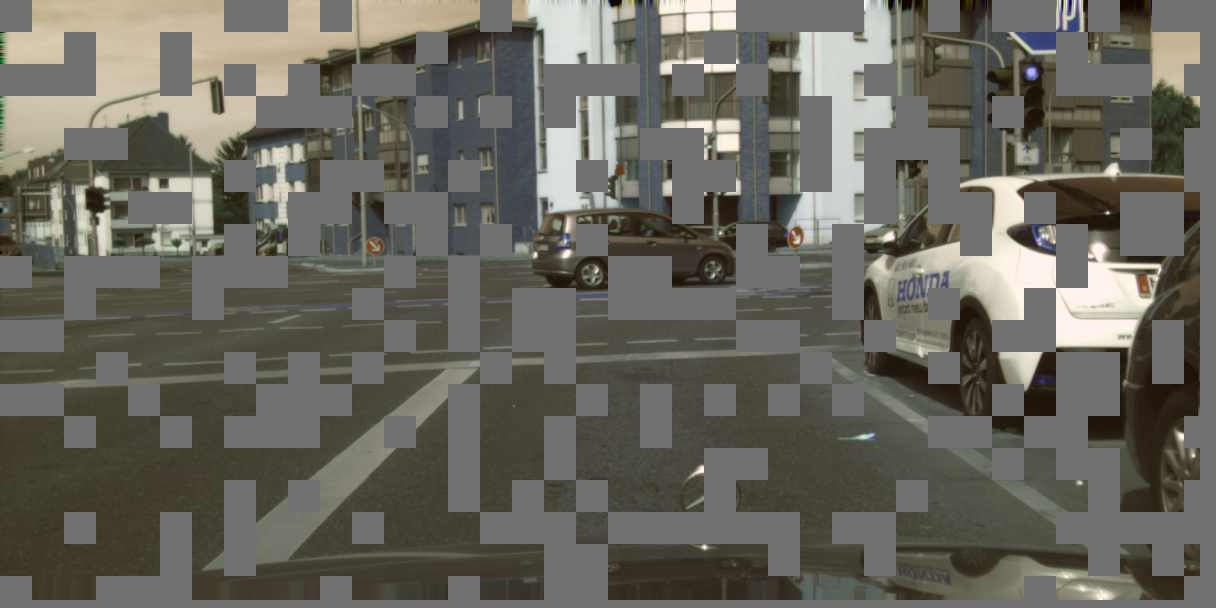}%
}
\subfigure[]{%
  \includegraphics[width=0.32\columnwidth,height=0.6in]{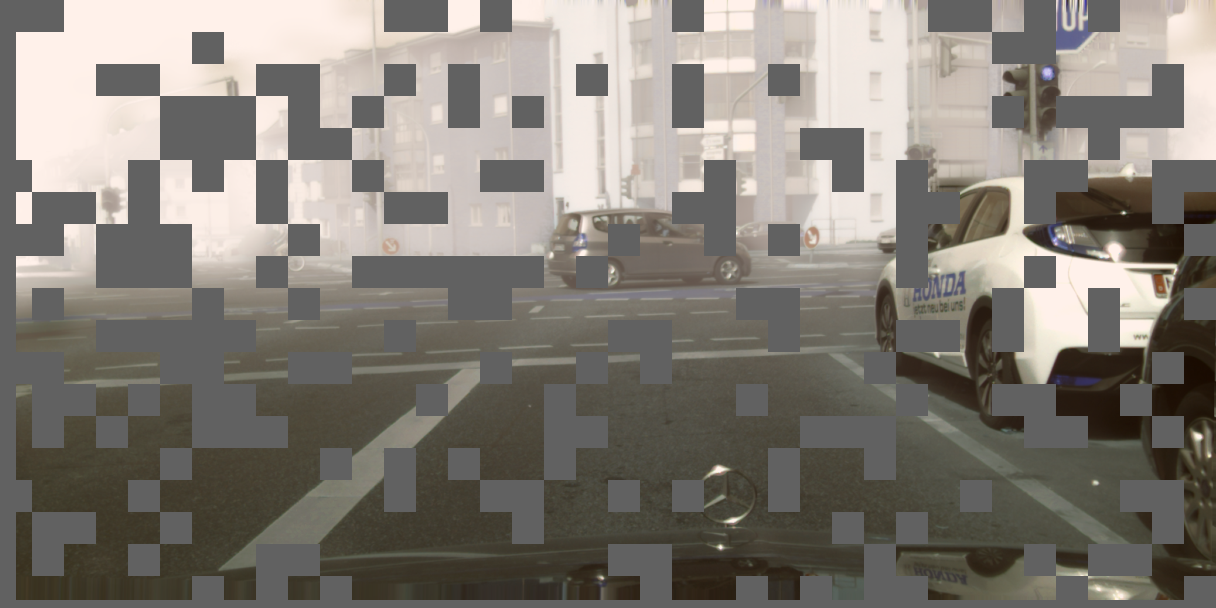}%
}
\subfigure[]{%
  \includegraphics[width=0.32\columnwidth, height=0.6in]{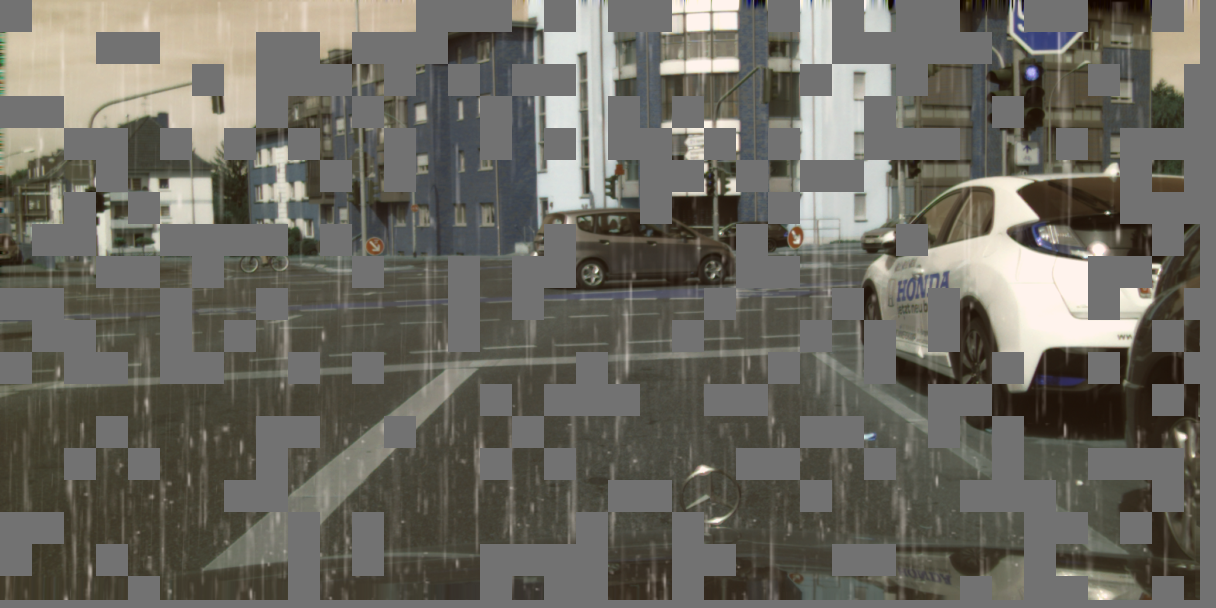}%
}
\caption{ Sample visualization of Dynamic Masking Process (DMP): (a) the masked original image from Cityscapes~\cite{cordts2016cityscapes}, (b) masked synthesized foggy image, (c) masked synthesized rainy image.}
\label{fig:DMP}
\end{figure}

Before feeding the input images into the CNN backbone, we implement our newly proposed Dynamic Masking Process (DMP) to generate masked images. This deep learning method can leverage contextual clues derived from surrounding image patches that may represent various parts of the object or its environment \cite{hoyer2023mic,hoyer2019grid,xie2022simmim}. In the training of deep learning models, the Dropout method is employed effectively by randomly ``dropping" neurons within the network to combat overfitting in CNNs \cite{srivastava2014dropout}. Drawing inspiration from \cite{xie2022simmim,hoyer2023mic}, and to bolster the learning of robust features while also curbing overfitting, our DMP selectively masks patches, each comprising 64 pixels across three input images. During model training, our DMP enhances robustness and prevents data-level overfitting by randomly ``dropping'' these patches (\textit{i.e.}, some specific pixel regions within the images), which is illustrated in Fig.~\ref{fig:DMP}. The patch mask rate is randomly sampled, following a uniform distribution ranging from 0 to 1.

\subsection{Image-level based Adaptation}
The image-level domain representation is derived from the feature extraction process of the backbone network, encompassing valuable global information such as style, scale, and illumination. 
These factors have the potential to greatly influence the performance of the object detection task~\cite{chen2018domain}.
To address this, we incorporate a domain classifier, which aims to classify the domains of the extracted image-level features and promote global alignment at the image level.
The domain classifier is implemented as two simple convolutional layers. It takes the image-level features as input and produces a prediction to identify the feature domain. A Binary Cross Entropy (BCE) loss is employed for the domain classifier as follows:

\begin{equation}\label{equ:img_loss}
 L_{img} = - \sum_{i=1}^{N}[G_{i} {\rm log} P_{i} + (1-G_{i}) {\rm log}(1-P_{i})], 
\end{equation}
where $i\in\{1,..., N\}$ represents the $N$ training images, 
the ground truth domain label of the $i$-th training image is denoted as $G_i\in\{1, 0\}$, where $G_i$ takes a value of 1 or 0 to represent the source and target domains, respectively. 
The prediction of the domain classifier for the $i$-th training image is denoted as $P_i$.

\subsection{Object-level based Adaptation} 
Besides the global differences at the image level, objects within different domains may exhibit variations in terms of appearance, size, color, and other characteristics. To address this, 
Each region proposal generated by the ROI Pooling layer is considered as a potential object of interest.
After obtaining the object-level domain representation via ROI pooling, we introduce an object-level domain classifier to discern the origin of the local features, which is implemented by three fully connected layers. The objective of the object-level domain classifier is to align the distribution of object-level features across different domains.
Similar to the image-level domain classifier, we utilize the BCE loss to train our object-level domain classifier:

\begin{equation}\label{equ:obj_loss}
 L_{obj} = - \sum_{i=1}^{N} \sum_{j=1}^{M}[G_{i,j} {\rm log} P_{i,j} + (1-G_{i,j}) {\rm log}(1-P_{i,j})], 
\end{equation}
where $j\in\{1,...,M\}$ is the $j$-th predicted object in the $i$-th image, $P_{i,j}$ is the prediction of the object-level domain classifier for the $j$-th region proposal in the $i$-th image, 
the corresponding binary ground-truth label for the source and target domains is denoted as $G_{i,j}$.

\subsection{Adversarial Gradient Reversal Layer}\label{Subsec:AdvGRL}

\begin{figure}[htb]
	\centering
	\includegraphics[width=0.95\columnwidth]{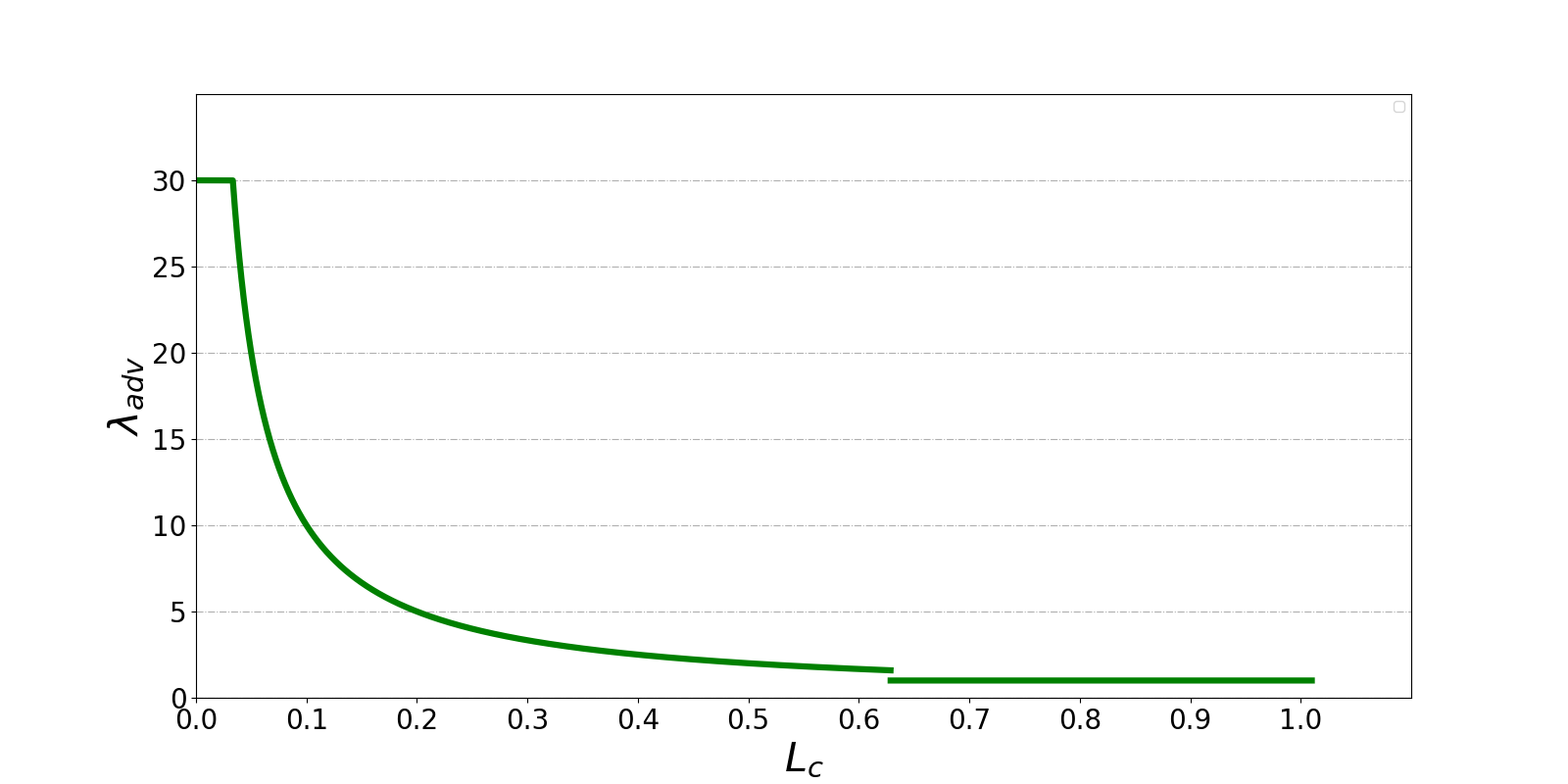}
	\caption{Illustration of the AdvGRL-based hard training example mining. We assign larger responses to harder training examples with lower domain classifier loss ($L_c$) values. In this paper, we set $\lambda_{0}=1$ and $\beta=30$.} 
	\label{fig:grl_loss} 
\end{figure}

In this section, we will begin by providing a brief overview of the original Gradient Reversal Layer (GRL), which serves as the foundation for our proposed Adversarial Gradient Reversal Layer (AdvGRL).
The original GRL was initially developed for unsupervised domain adaptation in image classification tasks~\cite{ganin2015unsupervised}. During forward propagation, the GRL leaves the input unchanged. However, during back-propagation, the gradient is reversed by multiplying it by a negative scalar before propagating it to the preceding layers of the base network. This reversal of gradients serves as a mechanism to confuse the domain classifier.  Like this, by reversing the gradient during back-propagation, the GRL encourages the base network to learn domain-invariant features, enabling DA.  The forward propagation of GRL is defined as: 

\begin{equation}\label{equ:grl_1}
  R_{\lambda}({\mathbf v}) = {\mathbf v},
\end{equation}
where ${\mathbf v}$ represents an input feature vector, and $R_{\lambda}$ represents the forward function performed by GRL. Back-propagation of GRL is defined as:

\begin{equation}\label{equ:grl_2}
  \frac{dR_{\lambda}}{d{\mathbf v}} = -\lambda \mathbf{I}, 
\end{equation}
where $\mathbf{I}$ is an identity matrix and $-\lambda$ is a negative scalar.
In the original GRL, a constant or varying value of $-\lambda$ is utilized, which is determined by the training iterations,
as described in~\cite{ganin2015unsupervised}. However, this approach overlooks the fact that different training samples may exhibit varying levels of challenge during transfer learning. To address this limitation, this paper introduces a novel AdvGRL that incorporates adversarial mining for hard examples.
This is achieved by replacing the parameter $\lambda$ with a new parameter $\lambda_{adv}$ in Eq.~(\ref{equ:grl_2}) of GRL, resulting in the proposed AdvGRL. Notably, the value of $\lambda_{adv}$ is determined as follows:

\begin{equation}\label{equ:ad_grl}
  \lambda_{adv} = \left \{
    \begin{aligned}
        & {\rm min} (\frac{\lambda_{0}}{L_{c}}, \beta), \qquad  & L_{c} < \alpha \\
        &\lambda_{0},   \qquad   & {\rm otherwise}, 
    \end{aligned}
    \right.
\end{equation}
where $L_{c}$ represents the loss of the domain classifier. $\alpha$ is a hardness threshold used to determine the difficulty level of the training sample. $\beta$ is the overflow threshold implemented to prevent the generation of excessive gradients during back-propagation. In our experiment, we set $\lambda_{0}=1$ as a fixed parameter.
Namely, when the domain classifier's loss $L_c$ is smaller, it indicates that the training sample's domain can be more easily identified by the classifier. In this case, the features associated with this sample are not the desired domain-invariant features, making it a more difficult example for domain adaptation. The relationship between $\lambda_{adv}$ and $L_c$ is visualized in Fig.~\ref{fig:grl_loss}.

Ous proposed AdvGRL serves two main purposes. 1) It utilizes the concept of gradient reversal during back-propagation to confuse the domain classifier, thereby promoting the generation of domain-invariant features. 2) AdvGRL enables adversarial mining for hard examples, meaning that it selectively focuses on challenging examples that contribute to the model's generalization.
Fig.~\ref{fig:da_faster_rcnn} illustrates the utilization of the AdvGRL in both the image-level and object-level DA modules.

\subsection{Domain-level Metric Learning based  Regularization}\label{Subsec:Metric Regularization}

A common transfer learning approach in many existing DA methods is to prioritize the transfer of features from a source domain $S$ to a target domain $T$.
Hence, they often overlook the potential advantages that a third-related domain can offer.
To explore the potential advantages of incorporating a third related domain,
we introduce an auxiliary domain for domain-level metric regularization that complements the source domain $S$. We leverage advanced data augmentation techniques to create this auxiliary domain $A$, which is particularly useful in autonomous driving scenarios where training data needs to be synthesized for different weather conditions based on existing clear-weather data.
As a result,  in our proposed architecture (as shown in Fig.~\ref{fig:da_faster_rcnn}), the source, auxiliary, and target domain images are regarded as aligned images, ensuring the enforcement of domain-level metric constraints across these three distinct domains.

The global image-level features of the $i$-th training image for the source ($S$), auxiliary ($A$), and target ($T$) domains are defined as $F_{i}^S$, $F_{i}^{A}$, and $F_{i}^{T}$, respectively. Our goal is to reduce the domain gap between $S$ and $T$ and ensure that the feature metric distance between $F_{i}^S$ and $F_{i}^{T}$ is closer compared to the distance between $F_{i}^S$ and $F_{i}^{A}$. This can be expressed as:

\begin{equation}\label{equ:reg_disance}
  d(F_{i}^{S}, F_{i}^{T}) < d(F_{i}^{S}, F_{i}^{A}),
\end{equation}    
where the metric distance between the corresponding features is denoted as $d(,)$. To implement this constraint, we can use a triplet structure where $F_{i}^S$, $F_{i}^{T}$, and $F_{i}^{A}$ are treated as the anchor, positive, and negative, respectively. Therefore, the image-level constraint in Eq.~(\ref{equ:reg_disance}) can be equivalently expressed as minimizing the image-level triplet loss:

\begin{equation}\label{equ:reg_loss_img}
  L^{R}_{img} = {\rm max}(d(F_{i}^{S}, F_{i}^{T} ) - d(F_{i}^{S}, F_{i}^{A}) + \delta , 0), 
\end{equation}
where the margin constraint is denoted as $\delta$, and in our experiments, $\delta$ is set as $1.0$.
Equivalently, the $i$-th training image's $j$-th object-level features of $S$, $A$, and $T$ are defined as $f_{i,j}^S$, $f_{i,j}^{A}$, and $f_{i,j}^{T}$ respectively. To apply our proposed metric regularization to the object-level features, we further minimize the object-level triplet loss:

\begin{equation}\label{equ:reg_loss_obj}
  L^{R}_{obj} = {\rm max}(d(f_{i,j}^{S}, f_{i,j}^{T} ) - d(f_{i,j}^{S}, f_{i,j}^{A}) + \delta, 0).  
\end{equation}

\subsection{Training Loss}
The overall training loss of the proposed network consists of several individual components. It can be expressed as follows:

\begin{equation}\label{equ:all_loss}
 L = \gamma*(L_{img} + L_{obj} + L^{R}_{img} + L^{R}_{obj}) + L_{cls} + L_{reg}, 
\end{equation}
where $L_{cls}$ and $L_{reg}$ represent the loss of classification and the loss of regression respectively. A weight parameter $\gamma$ is introduced to balance the Faster R-CNN loss and the domain adaptation loss, which is set as $0.1$. 
During the training phase, the network can be trained in an end-to-end manner utilizing a standard SGD algorithm.
During the testing phase, object detection can be performed using the Faster R-CNN  with the trained adapted weights.

\subsection{General Domain Adaptive Detection with Proposed Method} 
Our proposed method is designed to be versatile and adaptable to various domain adaptive object detection scenarios. Specifically, when dealing with scenarios where the target domain images are generated from the source domain with pixel-to-pixel correspondence, such as the Clear Cityscapes$\longrightarrow$Foggy Cityscapes, our method can be directly applied without any modifications.
To utilize our method with unaligned datasets in real-world scenarios,
where the target and source domains lack strict correspondence, such as the Cityscapes$\longrightarrow$KITTI, we can remove the $L^R_{obj}$ loss, which eliminates the requirement for object alignment during training. This allows our method to be applied directly without the need for object-level alignment.

\section{Experiments}~\label{Sec:Experiment}
\subsection{Benchmark}
\textbf{Cityscapes}~\cite{cordts2016cityscapes}: It is a widely used computer vision dataset that focuses on urban street scenes. There are 2,975 training sets and 500  validation sets from 27 different cities.
The dataset includes annotations for $8$ different categories. All images in the Cityscapes dataset are 3-channel $1024 \times 2048$ images.

\textbf{Foggy Cityscapes}~\cite{sakaridis2018semantic}: It is a public benchmark dataset created by simulating different intensity levels of fog on the original Cityscapes images. This dataset uses a depth map and a physical model~\cite{sakaridis2018semantic} to generate three levels of simulated fog. 

\begin{figure}[!t]
\centering
\subfigure[]{%
  \includegraphics[width=0.32\columnwidth, height=0.6in]{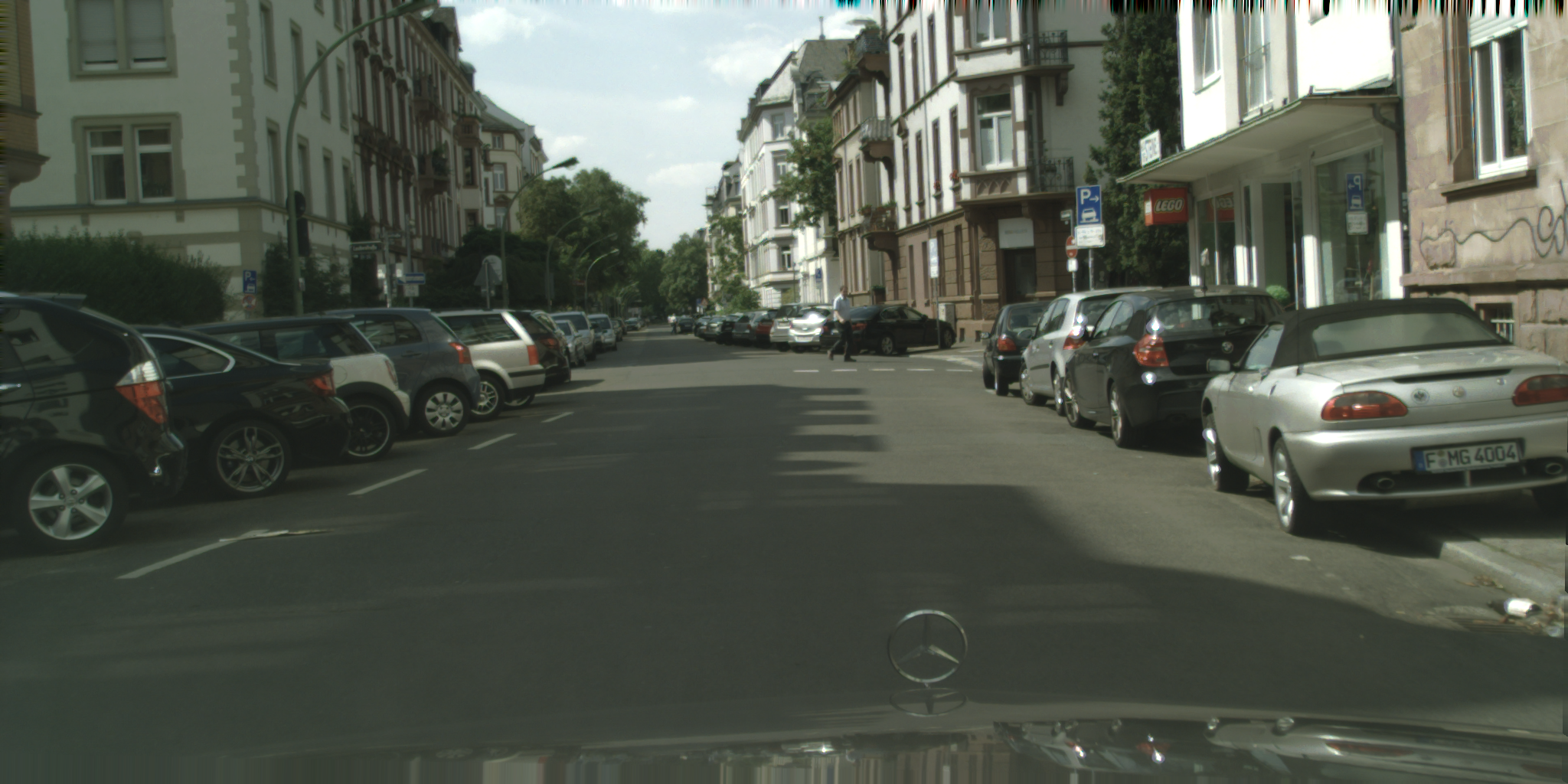}%
}
\subfigure[]{%
  \includegraphics[width=0.32\columnwidth,height=0.6in]{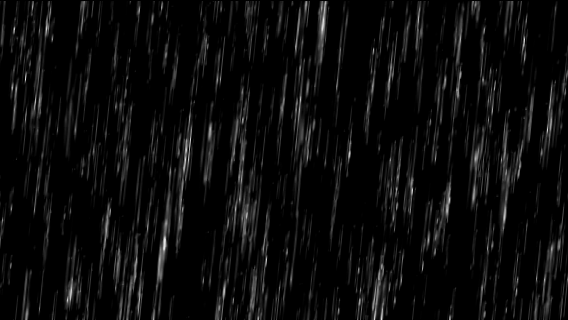}%
}
\subfigure[]{%
  \includegraphics[width=0.32\columnwidth, height=0.6in]{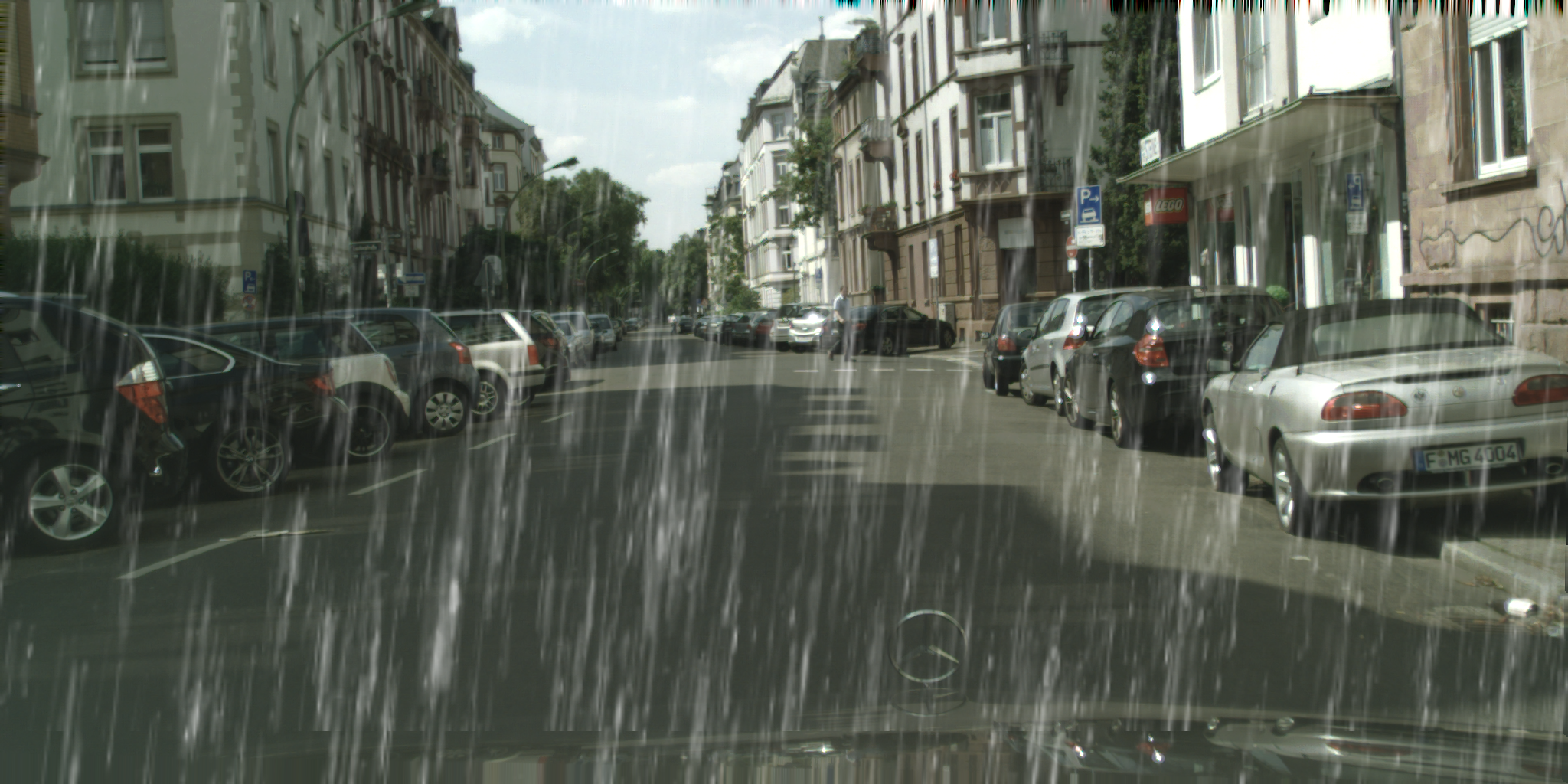}%
}
\caption{Illustration of synthesizing Rainy Cityscapes from the Cityscapes data: (a) the original image from Cityscapes~\cite{cordts2016cityscapes}, (b) rain map generated by  RainMix~\cite{guo2021efficientderain}, (c) synthesized rainy image.}
\label{fig:rain_domain}
\end{figure}

\textbf{Rainy Cityscapes}: We synthesize a rainy-weather dataset named as Rainy Cityscapes in this paper from the original Cityscapes dataset. Specifically, the training set of 3,475 images and the validation set of 500 images from Cityscapes are used to create the Rainy Cityscapes dataset by utilizing a novel data augmentation method called RainMix~\cite{guo2021efficientderain,hendrycks2020augmix}. 
To generate rainy Cityscapes images, we utilize a combination of techniques. First, we randomly sample a rain map from a publicly dataset of real rain streaks~\cite{garg2006photorealistic}. Next, we apply random transformations to the rain map using the RainMix technique. These transformations include rotation, zooming, translation, and shearing, which are randomly sampled and combined. 
Lastly, the rain maps after transformation are merged with the original source domain images,
resulting in the generation of rainy Cityscapes images. An example illustrating this process can be seen in Figure~\ref{fig:rain_domain}.

\begin{figure}[!t]
\centering
\subfigure[Small Fog]{%
  \includegraphics[width=0.32\columnwidth]{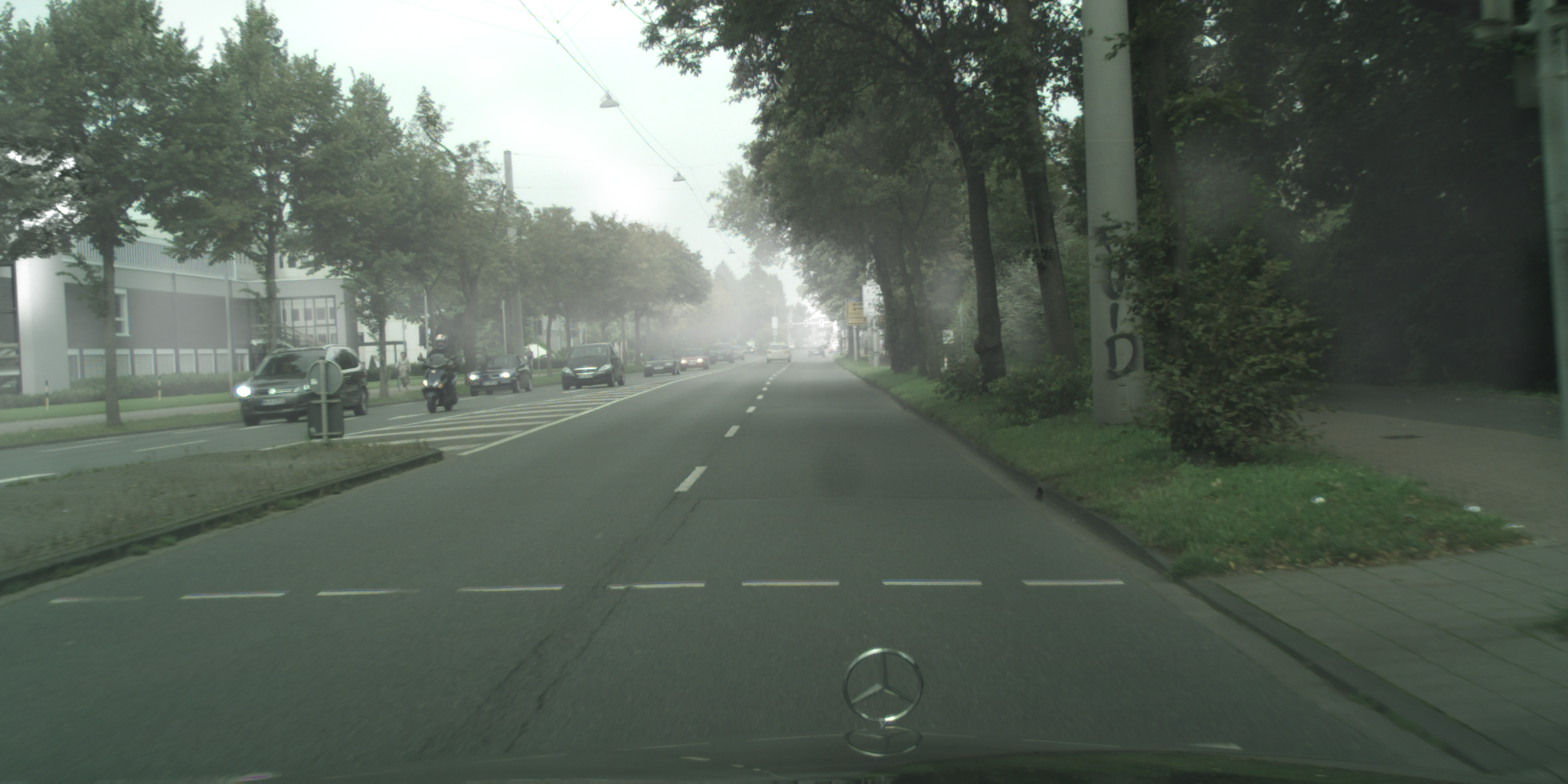}%
}
\subfigure[Medium Fog]{%
  \includegraphics[width=0.32\columnwidth]{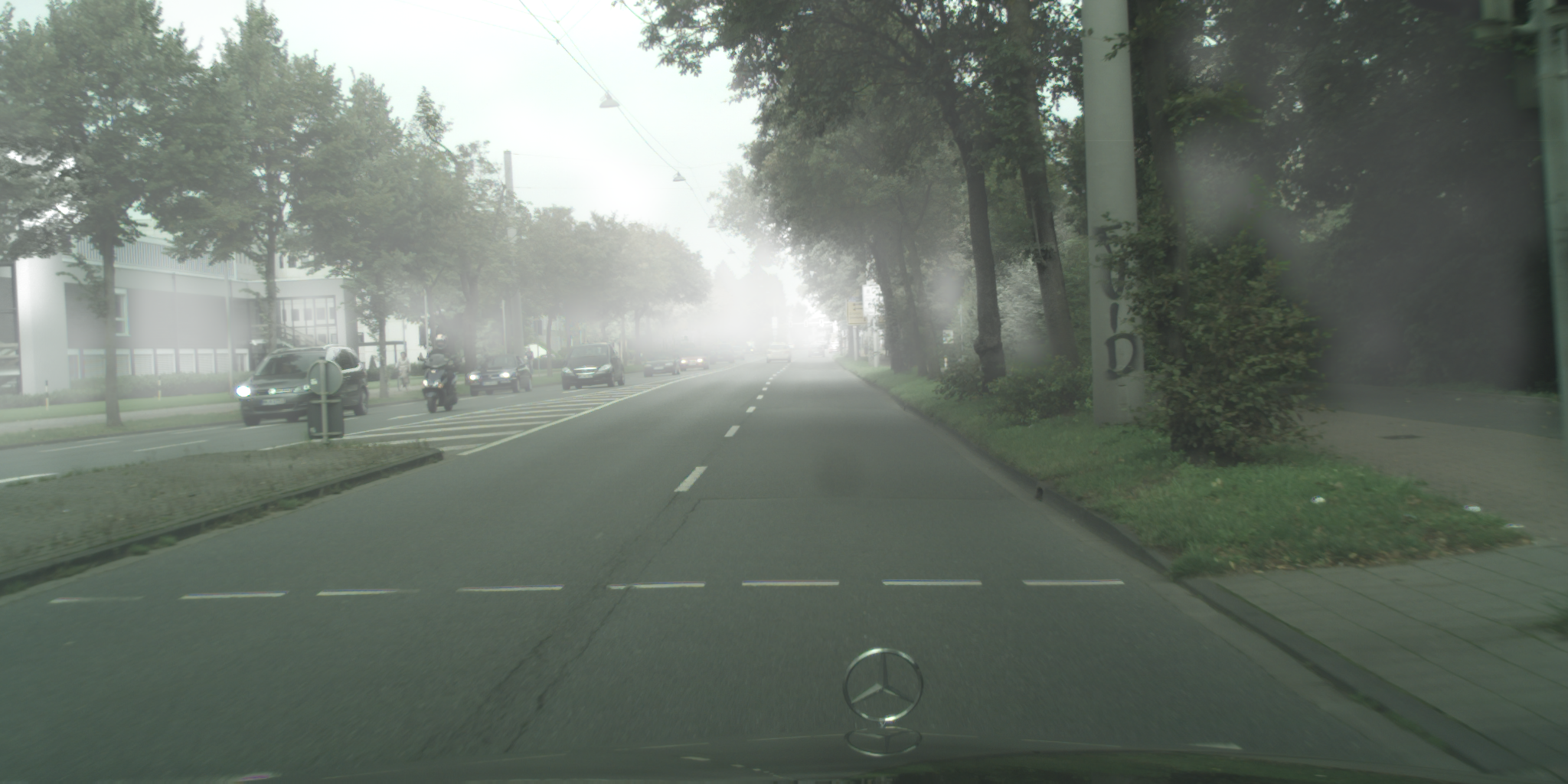}%
}
\subfigure[Large Fog]{%
  \includegraphics[width=0.32\columnwidth]{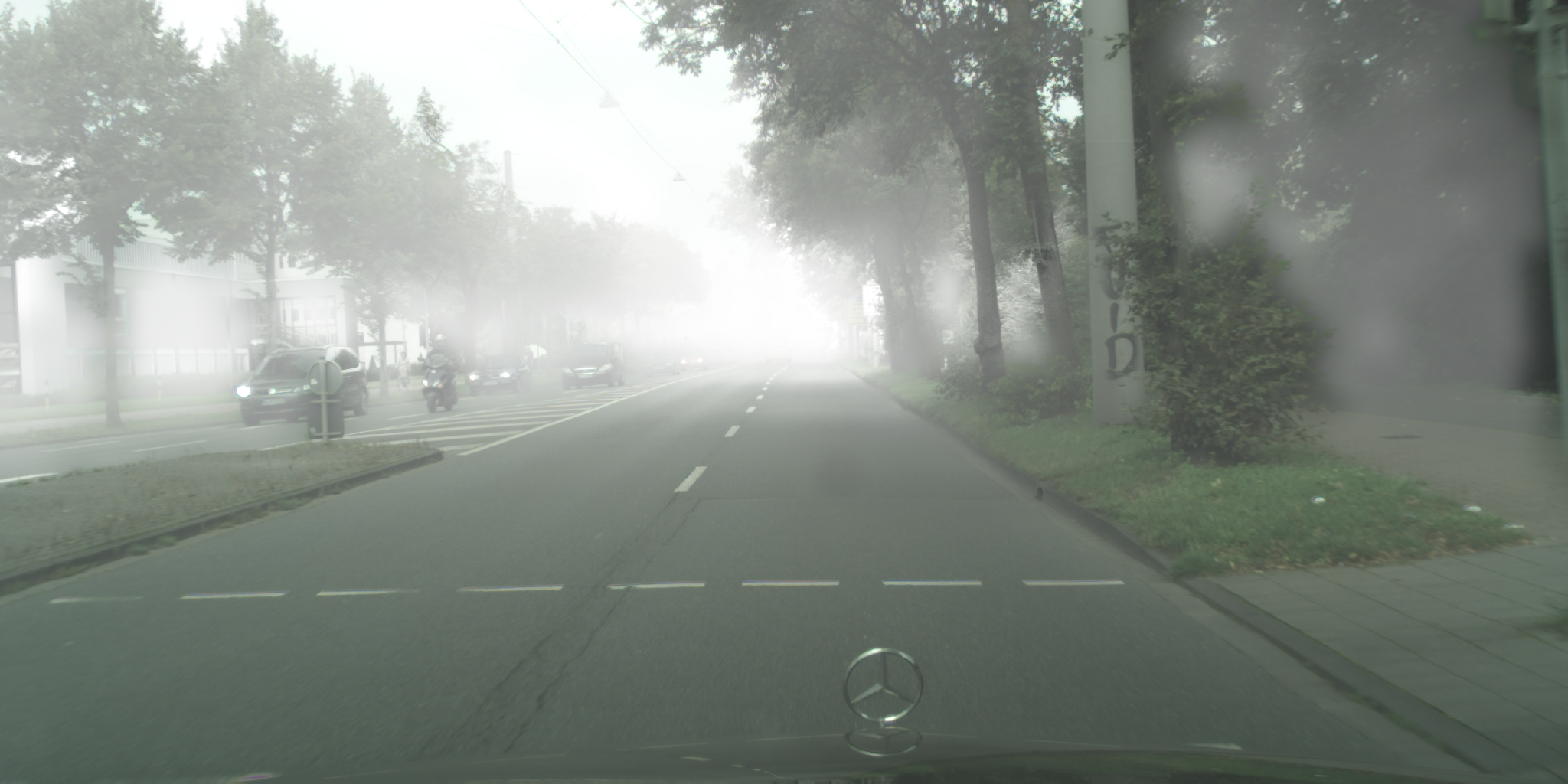}%
}
\subfigure[Small Rain]{%
  \includegraphics[width=0.32\columnwidth]{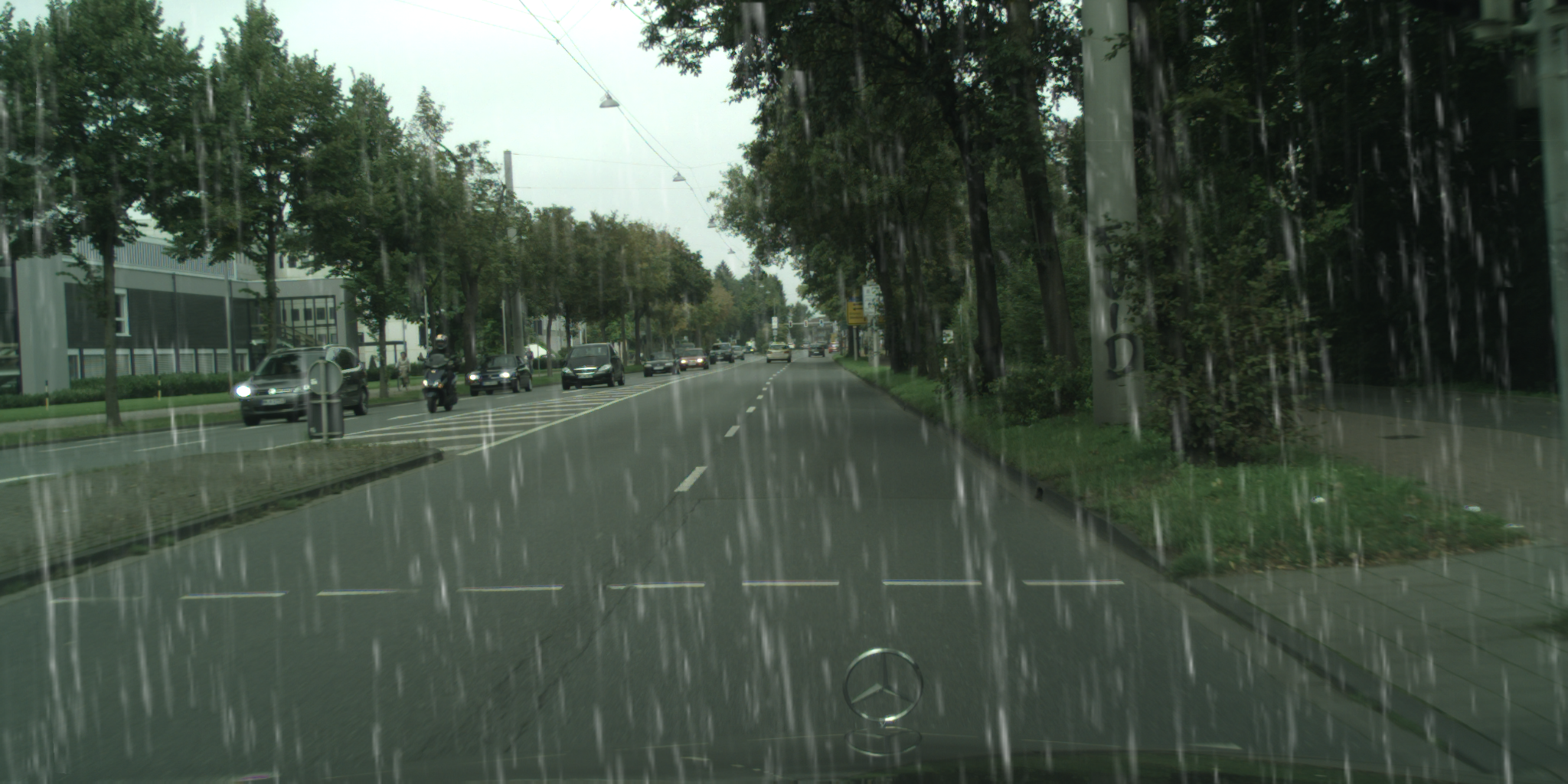}%
}
\subfigure[Medium Rain]{%
  \includegraphics[width=0.32\columnwidth]{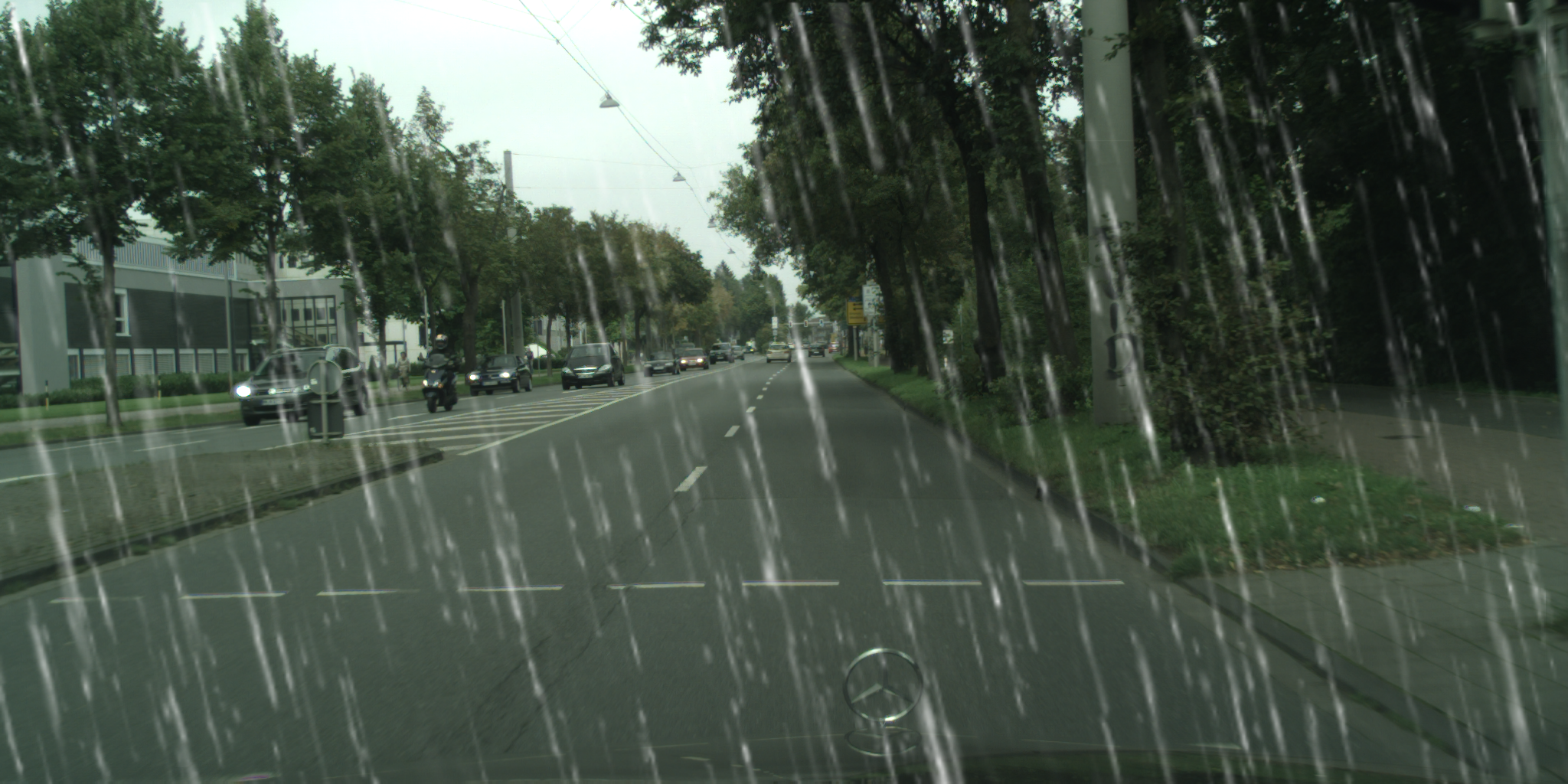}%
}
\subfigure[Large Rain]{%
  \includegraphics[width=0.32\columnwidth]{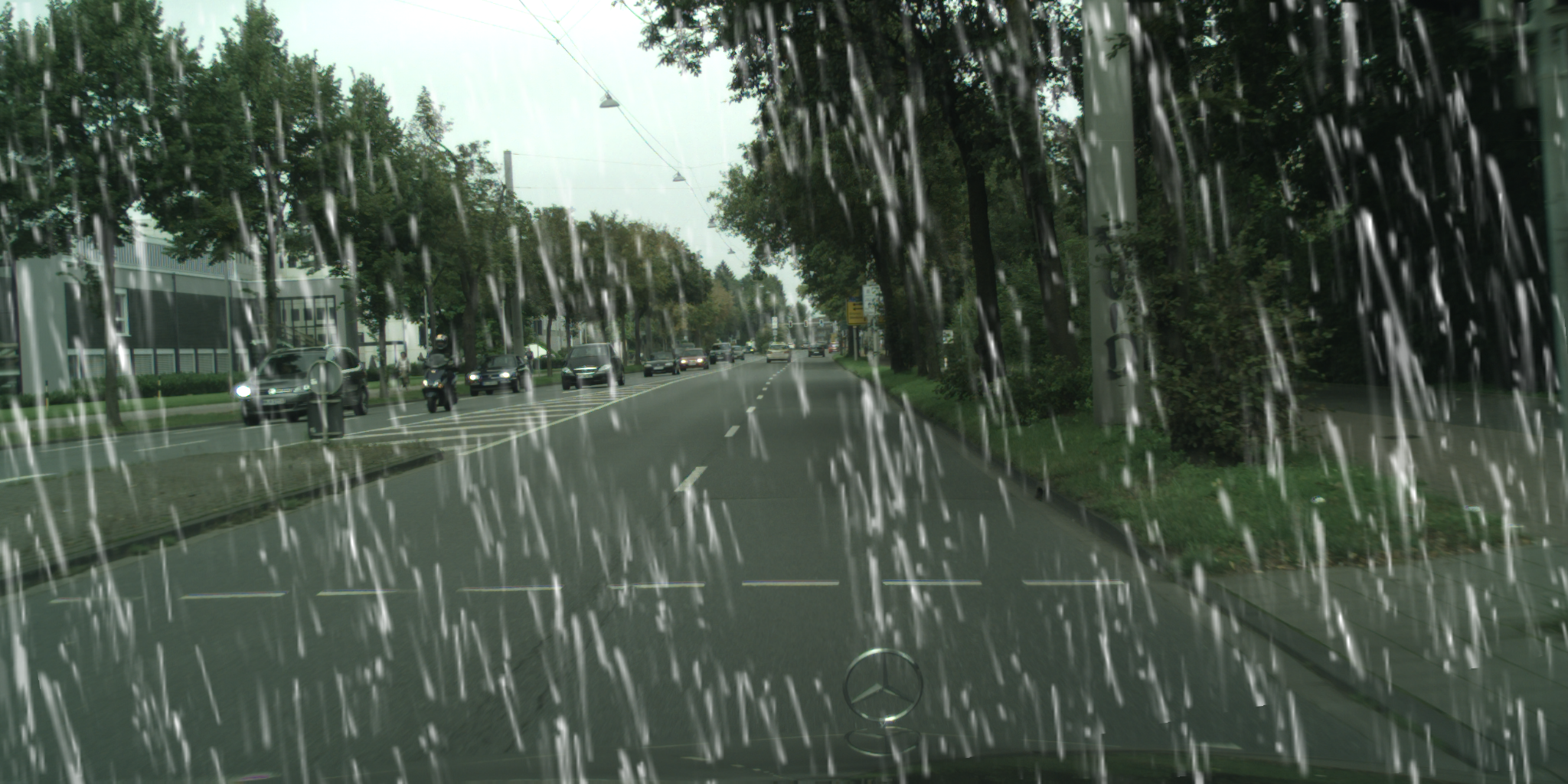}%
}
\caption{Sample visualization results for the Foggy Cityscapes and Rainy Cityscapes validation sets with different intensity levels.}
\label{fig:intensity}
\end{figure}

\textbf{Intensity levels of fog/rain}:  
For the Foggy Cityscapes and Rainy Cityscapes datasets, 
their number of images, resolution, and annotations are identical to those of the Clear  Cityscapes dataset.
Based on the physical model of~\cite{sakaridis2018semantic}, the different intensity levels of fog could be synthesized on the Foggy Cityscapes dataset. After obtaining the rain maps by  RainMix~\cite{guo2021efficientderain}, the intensity of rain maps could be further processed with different erosion levels. In these two ways, the different fog and rain levels (small, medium, large) can be synthesized, as shown in Fig.~\ref{fig:intensity}. Following the setting ~\cite{chen2018domain,xu2020cross,9439889}, the images with the highest intensity level of fog/rain are selected as the target domain for model training. The models trained with the highest intensity level will be then used to test the performance on the validation sets of different fog/rain intensity levels (small, medium, large).

\subsection{Experimental Setting}
\begin{table*}[]
\centering
\caption{Adaptation from Clear to Foggy:  Cityscapes$\rightarrow$Foggy Cityscapes experiment. Note that Oracle represents the Faster R-CNN trained on foggy Cityscape training set with all labels. The best performance is bold and the second best is underlined.}
\label{tab:foggy_comparison_result}
\resizebox{0.95\textwidth}{!}{%
\begin{tabular}{@{}cccccccccc@{}}
\toprule
\multicolumn{1}{c|}{Methodologies} &
  $C_{bus}$ &
  $C_{bicycle}$ &
  $C_{car}$ &
  $C_{mcycle}$ &
  $C_{person}$ &
  $C_{rider}$ &
  $C_{train}$ &
  \multicolumn{1}{c|}{$C_{truck}$} &
  mAP \\ \midrule
\multicolumn{1}{c|}{MCAR-ECCV'2020~\cite{zhao2020adaptive}} &
  44.1 &
  36.6 &
  43.9 &
  \textbf{37.4} &
  32.0 &
  42.1 &
  43.4 &
  \multicolumn{1}{c|}{31.3} &
  38.8 \\
\multicolumn{1}{c|}{MTOR-CVPR-2019~\cite{cai2019exploring}} &
  38.6 &
  35.6 &
  44.0 &
  28.3 &
  30.6 &
  41.4 &
  40.6 &
  \multicolumn{1}{c|}{21.9} &
  35.1 \\
\multicolumn{1}{c|}{DA-Faster-CVPR'2018~\cite{chen2018domain}} &
  49.8 &
  39.0 &
  53.0 &
  28.9 &
  35.7 &
  45.2 &
  45.4 &
  \multicolumn{1}{c|}{30.9} &
  41.0 \\
\multicolumn{1}{c|}{GPA-CVPR'2020~\cite{xu2020cross}} &
  45.7 &
  38.7 &
  54.1 &
  32.4 &
  32.9 &
  46.7 &
  41.1 &
  \multicolumn{1}{c|}{24.7} &
  39.5 \\
\multicolumn{1}{c|}{RPN-PR-CVPR'2021~\cite{zhang2021rpn}} &
  43.6 &
  36.8 &
  50.5 &
  29.7 &
  33.3 &
  45.6 &
  42.0 &
  \multicolumn{1}{c|}{30.4} &
  39.0 \\
\multicolumn{1}{c|}{UaDAN-TMM'2021~\cite{9439889}} &
  49.4 &
  38.9 &
  53.6 &
  32.3 &
  36.5 &
  46.1 &
  42.7 &
  \multicolumn{1}{c|}{28.9} &
  41.1 \\ 
\multicolumn{1}{c|}{HTCN-CVPR'2020~\cite{chen2020harmonizing}}      & 47.4 & 37.1 & 47.9 & 32.3 & 33.2 & 47.5 & 40.9 &  \multicolumn{1}{c|}{31.6} & 39.8 \\
\multicolumn{1}{c|}{SAPN-ECCV'2020~\cite{li2020spatial}}     & 46.8 & \underline{40.7} & 59.8 & 30.4 & 40.8 & 46.7 & 37.5 &  \multicolumn{1}{c|}{24.3} & 40.9 \\
\multicolumn{1}{c|}{MeGA-CDA-CVPR'2021~\cite{vibashan2021mega}}  & 49.2 & 39.0   & 52.4 &  \underline{34.5} &  37.7 & \underline{49.0}   & 46.9 &  \multicolumn{1}{c|}{25.4} & 41.8\\
\multicolumn{1}{c|}{UMT-CVPR2021~\cite{deng2021unbiased}}        & \textbf{56.6} & 37.3 & 48.6 & 30.4 & 33.0   & 46.7 & 46.8 &  \multicolumn{1}{c|}{\textbf{34.1}} & 41.7 \\
\multicolumn{1}{c|}{SCAN-AAAI'2022~\cite{li2022scan}}    & 48.6 & 37.3 & 57.3 & 31.0   & 41.7 & 43.9 & \underline{48.7} &  \multicolumn{1}{c|}{28.7} & 42.1 \\
\multicolumn{1}{c|}{ParaUDA-TITS’2022~\cite{zhang2022parauda}}  & 48.3 & 37.6 & 52.5 &  33.5 & 36.7 & 46.7 & 45.9 &  \multicolumn{1}{c|}{ \underline{32.3}} & 41.7 \\
\multicolumn{1}{c|}{ConfMix-WACV'2023~\cite{mattolin2023confmix}}  & 45.8 & 33.5 & \textbf{62.6} & 28.6 & \textbf{45.0}   & 43.4 & 40.0   &  \multicolumn{1}{c|}{27.3} & 40.8 \\
\multicolumn{1}{c|}{SDAYOLO-TIV'2023~\cite{li2022cross}}   & 40.5 & 37.3 &  \underline{61.9} & 24.4 &  \underline{42.6} & 42.1 & 39.5 &  \multicolumn{1}{c|}{23.5} & 39.0   \\
\multicolumn{1}{c|}{MS-DAYOLO-TIP'2023~\cite{10070607}}    & 51.0   & 36.0   & 56.5 & 27.5 & 39.6 & 46.5 &  45.9 &  \multicolumn{1}{c|}{28.9} & 41.5  \\

  \midrule
\multicolumn{1}{c|}{Ours  w/o Auxiliary Domain} &
  48.4 &
  36.7 &
  53.5 &
  26.1 &
  36.1 &
  45.9 &
  39.1 &
  \multicolumn{1}{c|}{29.3} &
  40.2 \\
\multicolumn{1}{c|}{Ours} &
  \underline{51.2} &
  39.1 &
  54.3 &
  31.6 &
  36.5 &
  46.7 &
  \underline{48.7} &
  \multicolumn{1}{c|}{30.3} &
  \underline{42.3} \\
\multicolumn{1}{c|}{Ours$\textbf{+}$} &
  48.7 &
  \textbf{41.6} &
  55.8 &
  33.3 &
  36.5 &
  \textbf{49.1} &
  \textbf{51.3} &
  \multicolumn{1}{c|}{30.0} &
  \textbf{43.4} \\ 
\midrule
\multicolumn{1}{c|}{Oracle} &
  49.9 &
  45.8 &
  65.2 &
  39.6 &
  46.5 &
  51.3 &
  34.2 &
   \multicolumn{1}{c|}{32.6} &
  45.6 \\ \bottomrule
\end{tabular}%
}
\end{table*}

\begin{table*}[]
\centering
\caption{Adaptation from Clear to Rainy:   Cityscapes$\rightarrow$Rainy Cityscapes experiment. Note that Oracle represents the Faster R-CNN trained on rainy Cityscapes training set with all labels. The best performance is bold and the second best is underlined.}
\label{tab:rainy_comparison_result}
\resizebox{0.95\textwidth}{!}{%
\begin{tabular}{@{}cccccccccc@{}}
\toprule
\multicolumn{1}{c|}{Methodologies} &
  $C_{bus}$ &
  $C_{bicycle}$ &
  $C_{car}$ &
  $C_{mcycle}$ &
  $C_{person}$ &
  $C_{rider}$ &
  $C_{train}$ &
  \multicolumn{1}{c|}{$C_{truck}$} &
  mAP \\ \midrule
\multicolumn{1}{c|}{Faster R-CNN (source only)}  & 46.3 & 26.0 & 54.8 &  25.8 & 34.7 & 35.9 & 26.9 &  \multicolumn{1}{c|}{23.9} & 34.3 \\
\multicolumn{1}{c|}{DA-Faster-CVPR'2018~\cite{chen2018domain}}  & 54.6 & 33.8 & 59.6 & 32.4 & 38.2  & 41.6 & 42.9   &  \multicolumn{1}{c|}{33.8} & 42.1 \\
\multicolumn{1}{c|}{MS-DAYOLO-TIP'2023~\cite{10070607}}    &\textbf{60.3}   &  34.4    & \textbf{67.7}  &31.9  &\textbf{47.2}  & \textbf{47.4}  &31.9  &  \multicolumn{1}{c|}{30.7 } & 44.0   \\
  
  \midrule
\multicolumn{1}{c|}{Ours  w/o Auxiliary Domain} &
  55.5 &
  \underline{35.3} &
  60.0 &
   32.8 &
  38.3 &
  22.1 &
   \underline{49.3} &
  \multicolumn{1}{c|}{33.1} &
  43.4 \\
\multicolumn{1}{c|}{Ours} &
   \underline{60.0} &
    35.3 &
   60.6 &
  \underline{33.8} &
   38.8 &
   42.9 &
  \textbf{52.4} &
  \multicolumn{1}{c|}{\underline{36.3}} &
  \underline{45.0} \\ 
\multicolumn{1}{c|}{Ours$\textbf{+}$} &
   59.7 &
  \textbf{39.0} &
   \underline{61.7} &
  \textbf{34.4} &
   \underline{40.2} &
   \underline{47.0} &
  47.2 &
  \multicolumn{1}{c|}{\textbf{38.8}} &
  \textbf{46.0} \\ 
  \midrule
\multicolumn{1}{c|}{Oracle} &
  58.5 &
  35.3 &
  66.9 &
  35.7 &
  47.5 &
  50.9 &
  37.4 &
   \multicolumn{1}{c|}{40.5} &
  46.6 \\ \bottomrule
\end{tabular}%
}
\end{table*}

\noindent \textbf{Dataset setting}: 
We conducted two main experiments in this paper: 1) Clear to Foggy Adaptation, denoted as Clear Cityscapes$\rightarrow$Foggy Cityscapes, the labeled training set of Clear Cityscapes~\cite{cordts2016cityscapes} and the unlabeled training set of Foggy Cityscapes~\cite{sakaridis2018semantic} are used as the source and target domains during training, respectively. Subsequently, the trained model was evaluated by the Foggy Cityscapes validation set to report the performance. Rainy Cityscapes training set is used as the Auxiliary Domain $A$ in this Clear to Foggy Adaptation experiment.  
2) Clear to Rainy Adaptation, denoted as the Clear Cityscapes$\rightarrow$Rainy Cityscapes, where the labeled training set of Clear Cityscapes~\cite{cordts2016cityscapes} and the unlabeled training set of Rainy Cityscapes are used as the source and target domains during training, respectively. Then the trained model was evaluated on Rainy Cityscapes validation set to report the performance. Foggy Cityscapes training set is used as the Auxiliary Domain $A$ in this Clear to Rainy Adaptation experiment. Additionally, we analyzed the transfer learning performance on different intensity levels of fog and rain (small, medium, and large).

\noindent \textbf{Training setting}: We utilize ResNet-50 as the backbone for the Faster R-CNN~\cite{ren2015faster}. Following in~\cite{chen2018domain,ren2015faster}, during training, We utilize back-propagation and stochastic gradient descent (SGD) to optimize all the deep learning methods in our approach.
The initial learning rate of $0.01$ for $50,000$ iterations is used in all model training. Afterward, the learning rate is reduced to $0.001$ and training continues for an additional $20,000$ iterations.
Weight decay is set as $0.0005$ and momentum is set as $0.9$ for all experiments. Each training batch consists of three images from the source, target, and auxiliary domains respectively.
For comparison purposes, we set the $\lambda$ value in the original GRL (Equation (\ref{equ:grl_2})) to $1$. In the AdvGRL (Equation (\ref{equ:ad_grl})), the hardness threshold $\alpha$ is set to $0.63$, which is computed by averaging the parameters
in Equation (\ref{equ:img_loss}) with setting ($P_i=0.7, G_i=1$ and $P_i=0.3, G_i=0$).
In the subsequent analysis, we refer to ``Reg + AdvGRL" as our proposed DA method. Additionally, ``Reg + AdvGRL + DMP" is designated as our enhanced DA method, termed DA+, which is named ``Ours+" for short in the tables.

\noindent \textbf{Evaluation metrics}: We calculate the Average Precision for each category and the mean Average Precision across all categories using an Intersection over the Union threshold of $0.5$.

\subsection{Adaptation from Clear to Foggy}

Table~\ref{tab:foggy_comparison_result} presents the results of our experiments on weather adaptation from clear to foggy. 
In comparison to other DA methods, our proposed DA+ method achieves the highest performance on Foggy Cityscape, with a mAP of $43.4\%$, which outperforms the third-best method SCAN~\cite{li2022scan} by a margin of $1.3\%$ in terms of mAP improvement.
The proposed DA+ method effectively reduces the domain gap across various categories,
\textit{e.g.}, the rider got $49.1\%$ and the bicycle got $41.6\%$ as the second best performance, and the train got $51.3\%$ as the best performance in AP, which is the highlight in Table~\ref{tab:foggy_comparison_result}.
While UMT got $56.6\%$ in the bus and $34.1\%$ in the truck, ConfMix got $62.6\%$ in the car, MeGA-CDA got $49.0\%$ in the rider,
our proposed DA+ method exhibits similar performance across them with only minor differences. However, our proposed DA+ method achieves the highest overall mAP detection performance on Foggy Cityscapes among the recent DA methods.

\begin{table}[]
\centering
\caption{Ablation study of components on the experiments of Cityscapes$\rightarrow$Foggy Cityscapes and  Cityscapes$\rightarrow$Rainy Cityscapes.}
\label{tab:Ablation}
\resizebox{\columnwidth}{!}{%
\begin{tabular}{@{}c|ccccc|cc@{}}
\toprule
Methods                &Img  & Obj       & AdvGRL    & Reg     & DMP    & Foggy mAP           & Rainy mAP   \\ \midrule
Source only            &              &           &           &    &         & 23.41      &34.35     \\
\textbf{w/} DMP     &     &      &           &       & \checkmark      & 24.54    &35.90      \\
img \textbf{w/} GRL                & \checkmark    &           &      &       &           & 38.10     &36.41      \\
obj \textbf{w/} GRL                &              & \checkmark &       &      &           & 38.02     &37.92      \\
img+obj \textbf{w/} GRL (Baseline) & \checkmark    & \checkmark &      &       &           & 38.43    &41.02       \\
img+obj \textbf{w/}AdvGRL         & \checkmark    & \checkmark & \checkmark &     &        & 40.23   &43.44        \\
img+obj+ Reg \textbf{w/} GRL       & \checkmark    & \checkmark &           & \checkmark   &   & 41.97   &44.44        \\
img+obj+Reg \textbf{w/} AdvGRL    & \checkmark    & \checkmark & \checkmark & \checkmark  &   &42.34 & 45.07 \\ 
img+obj+Reg+DMP \textbf{w/} AdvGRL    & \checkmark    & \checkmark & \checkmark & \checkmark  & \checkmark  & \textbf{43.42} & \textbf{46.04} \\ 

\bottomrule
\end{tabular}%
}
\end{table}
\vspace{-1em}

\begin{figure}[htb]
	\begin{minipage}[b]{0.5\textwidth}
		\centering
		\includegraphics[width=0.94\textwidth]{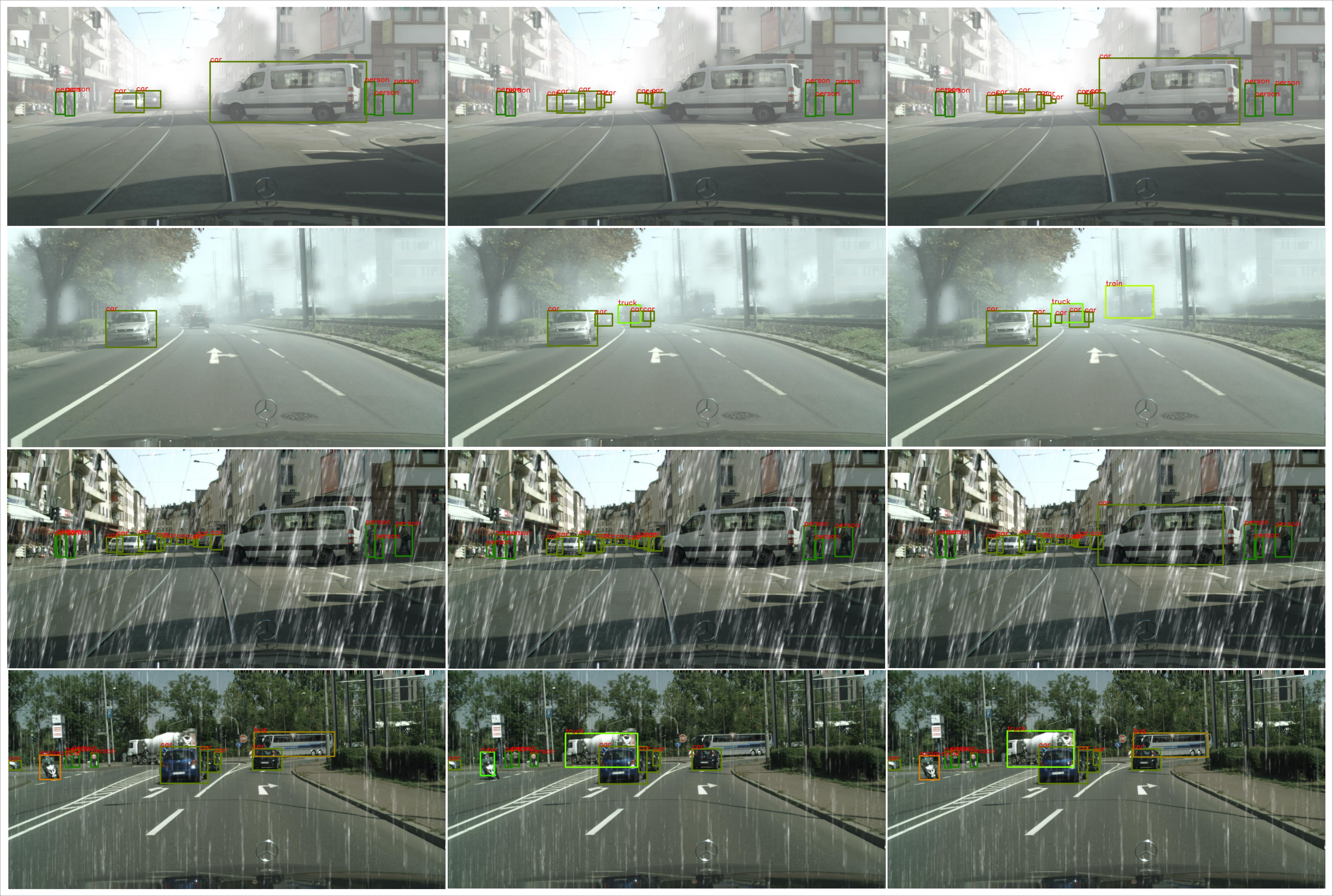}
	\end{minipage}
\caption{Visualization of the qualitative detection on Foggy/Rainy Cityscapes (validation sets). First column (Source only): original Faster R-CNN \textbf{w/} DA, Second column (Baseline): Faster R-CNN with image-level and object-level adaptations \textbf{w/} GRL, Third column: Proposed DA Method. Top: foggy weather, Bottom: rainy weather.}  
\label{fig:visual_detection} 
\end{figure}

\subsection{Adaptation from Clear to Rainy}
In the Clear to Rainy adaptation, the only difference during training is the exchange of domains, where the unlabelled Rainy Cityscapes training set serves as the target domain, while the Foggy Cityscapes training set is used as the auxiliary domain. 
Table~\ref{tab:rainy_comparison_result} presents the results of domain adaptation from clear to rainy weather.
Due to the page limit, we choose the methods with the publicly available source code which perform very well in the Clear to Foggy Adaptation experiment as the comparison methods in this Clear to Rainy Adaptation experiment, \textit{i.e.}, DA-Faster~\cite{chen2018domain}, MS-DAYOLO~\cite{10070607}. 
Similar to the Clear to Foggy Adaptation, our proposed DA+ method got the best overall mAP (46.0\%) detection performance on Rainy Cityscapes compared to the comparison methods.

\subsection{Ablation Study of Components}
We conduct an analysis of the individual proposed components of our DA object detection method.  The experiments are conducted on the Cityscapes$\rightarrow$Foggy Cityscapes and Cityscapes$\rightarrow$Rainy Cityscapes tasks, using the ResNet-50 backbone. The results of the ablation study are presented in Table~\ref{tab:Ablation}.
In the first row of the table, image-level and object-level adaptation modules are labels as `img' and `obj',  respectively. `AdvGRL' and `Reg' indicate the proposed Adversarial GRL and domain-level metric regularization, respectively.
The `img+obj+GRL' configuration represents the \textit{Baseline} model used in our experiments. We also evaluate two additional configurations: `img+obj+AdvGRL' and `img+obj+AdvGRL+Reg'. Additionally, we include the `Source only' configuration, which refers to the Faster R-CNN model trained solely on labeled source domain images without any DA methods.
The ablation study presented in Table~\ref{tab:Ablation} provides clear evidence of the positive impact of each proposed component in the DA method for both foggy and rainy weather scenarios.  
Furthermore, we provide qualitative visualization of the object detection results in Fig.~\ref{fig:visual_detection}.

\begin{figure*}[!t]
\centering
\subfigure[]{%
  \includegraphics[width=0.49\columnwidth,height=1.6in]{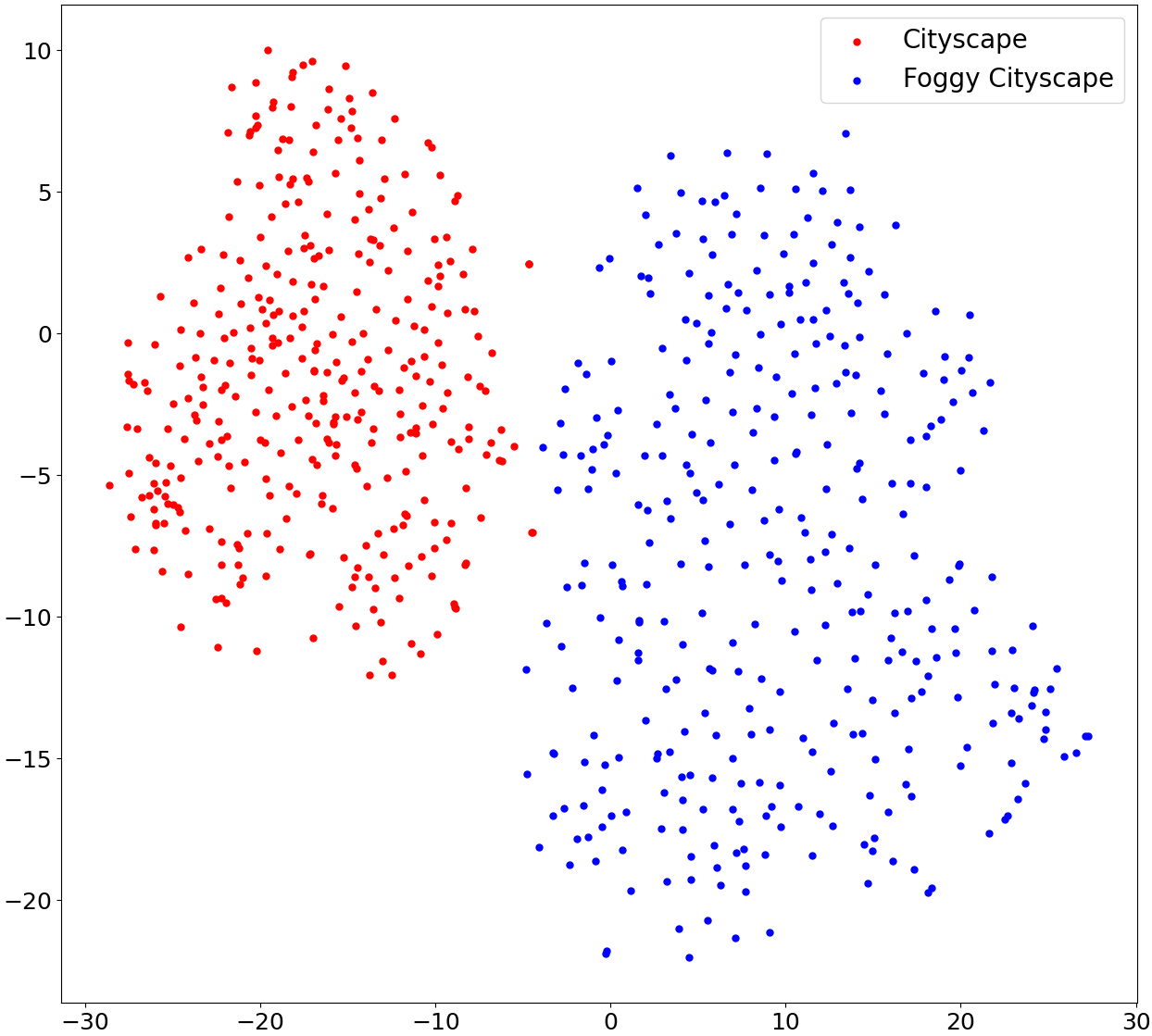}%
}
\subfigure[]{%
  \includegraphics[width=0.49\columnwidth,height=1.6in]{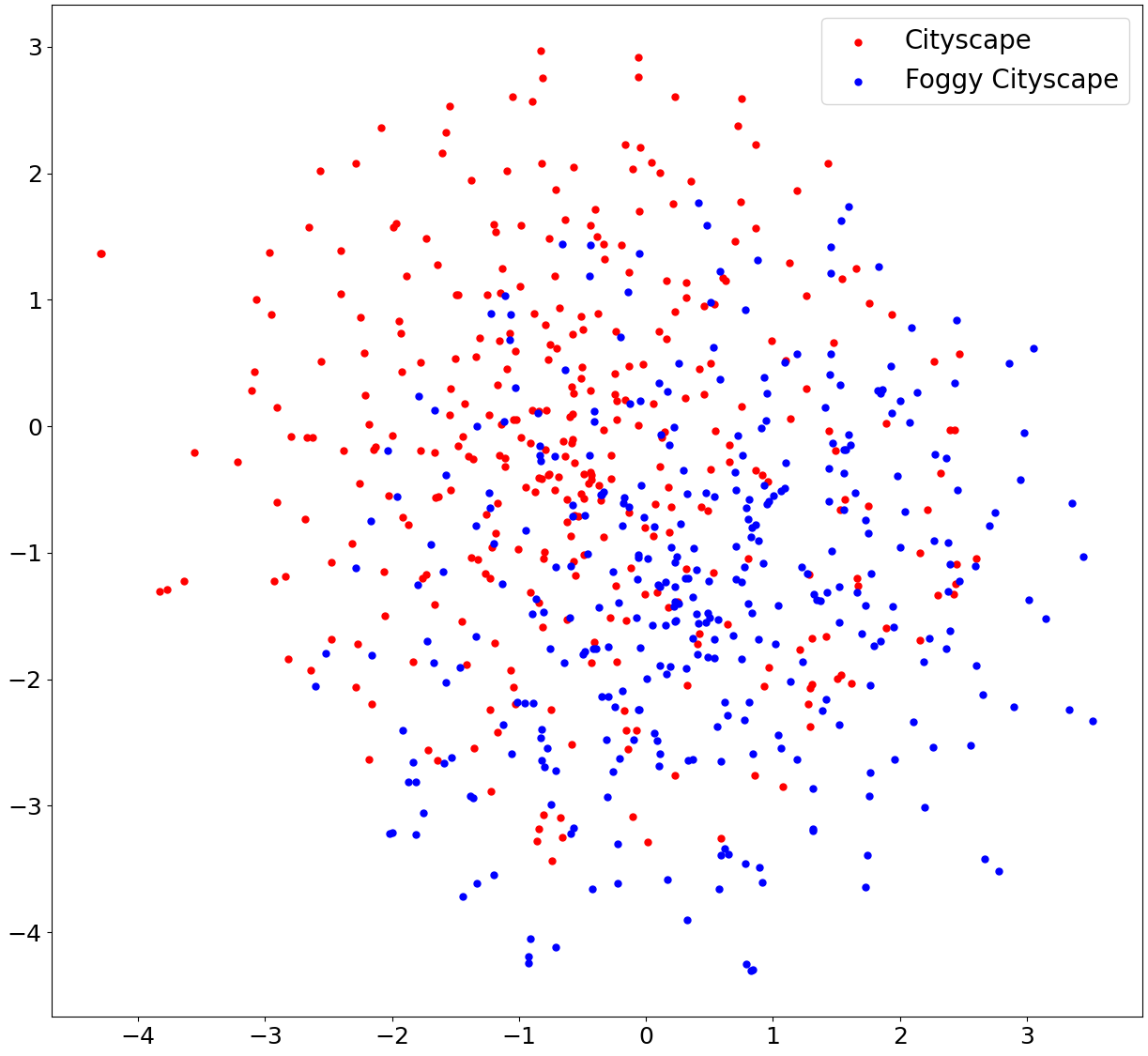}%
}
\subfigure[]{%
  \includegraphics[width=0.49\columnwidth,height=1.6in]{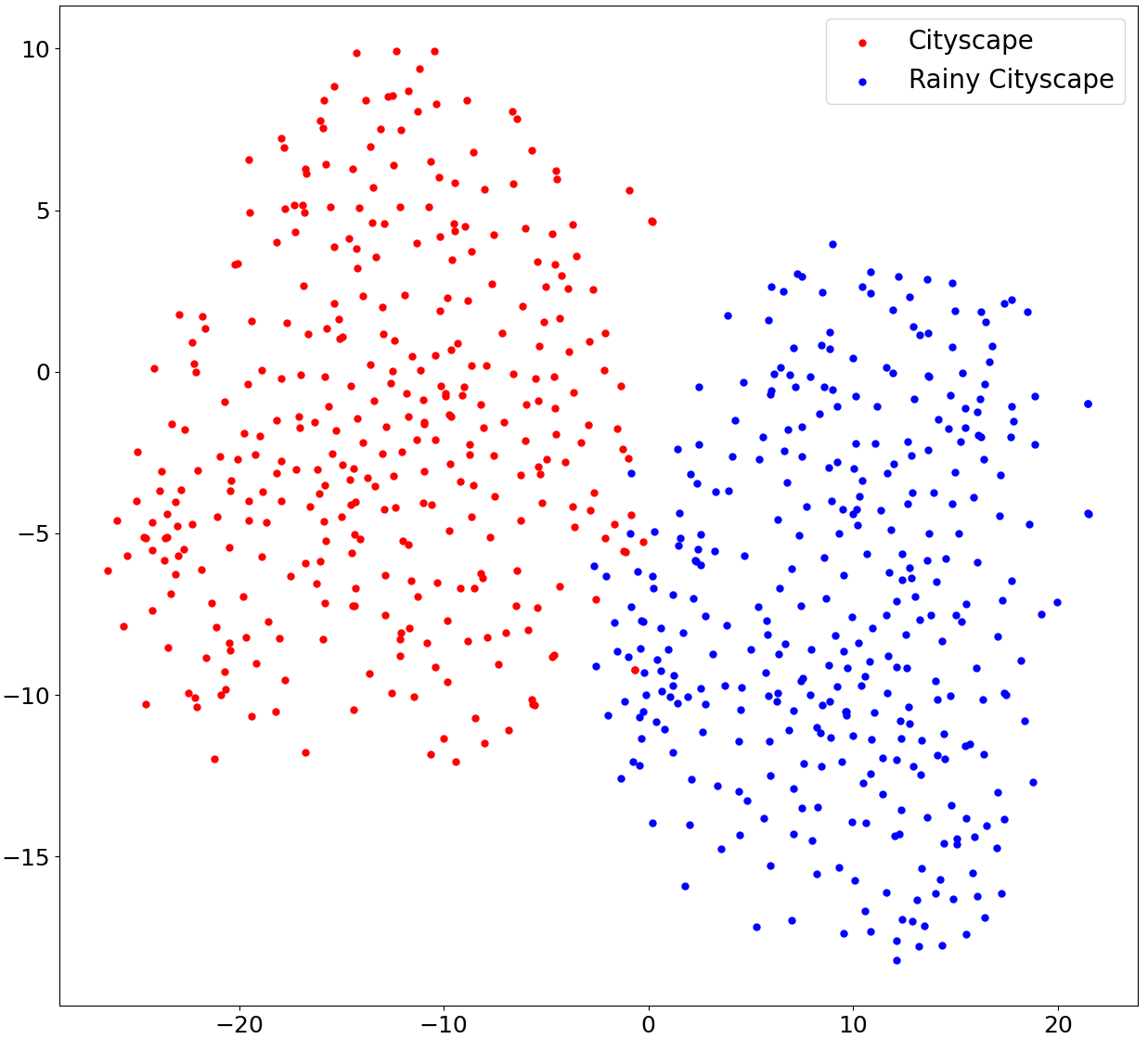}
}
\subfigure[]{%
  \includegraphics[width=0.49\columnwidth,height=1.6in]{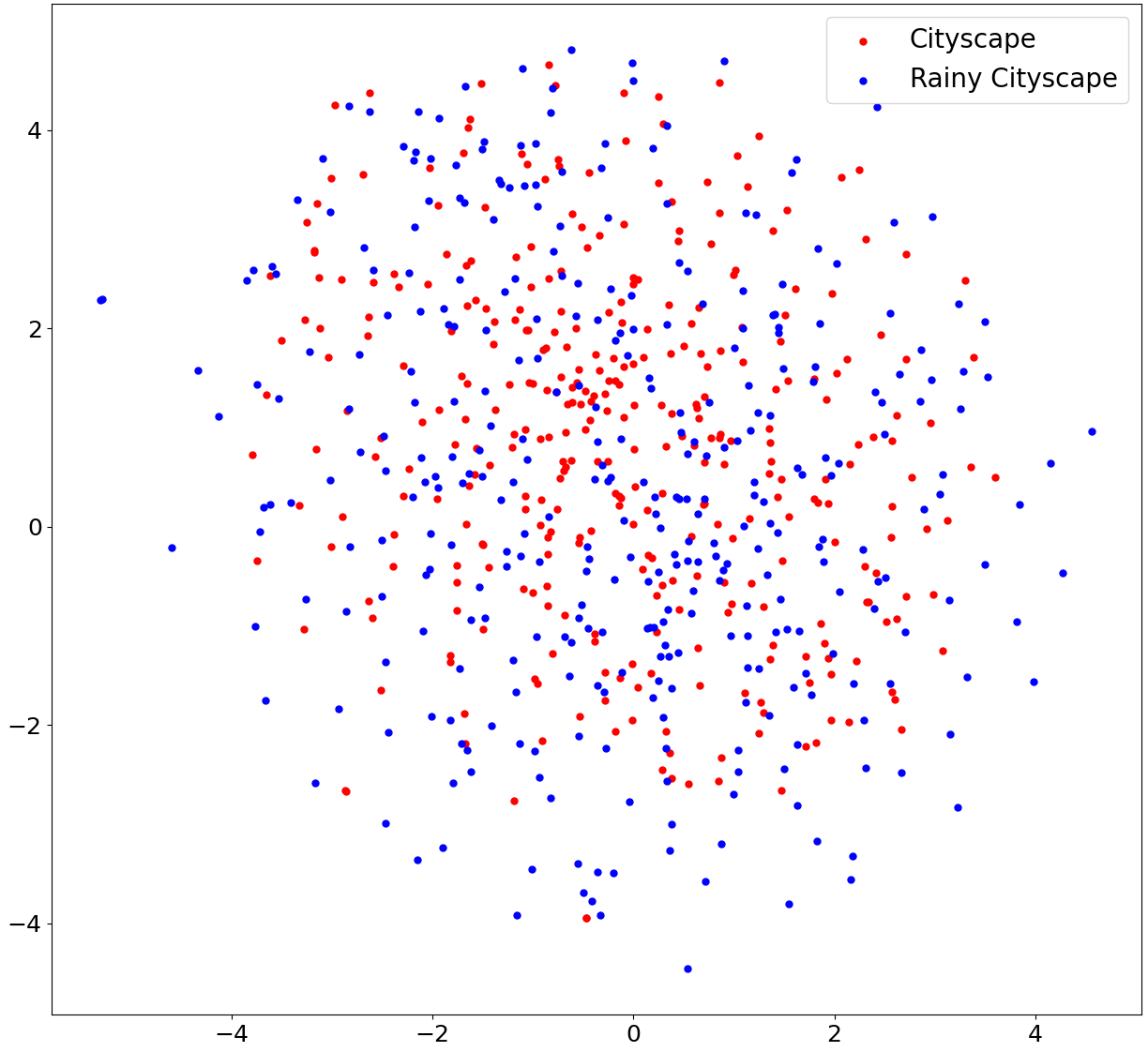}%
}
\caption{Feature distribution visualization by 
t-SNE~\cite{van2008visualizing} before and after domain adaptation. Clear to Foggy Adaptation: (a) original distribution before adaptation, (b) aligned distribution after the proposed adaptation. Clear to Rainy Adaptation: (c) original distribution before adaptation, (d) aligned distribution after the proposed adaptation. It is recommended to view this figure in color.}
\label{fig:feature_distribution}
\end{figure*}

\subsection{Adaptation of Different-intensity Fog and Rain} 

Moreover, both simulated Foggy and Rainy Cityscapes datasets contain three levels of intensity, namely Small, Medium, and Large as depicted in Fig.~\ref{fig:intensity}. Following the previous  works~\cite{chen2018domain,xu2020cross,9439889}, we only utilize the Large intensity level as the target domain during training for both fog and rain. After training, the trained models on the validation set of Rainy Cityscapes and  Foggy Cityscapes with images of different intensity levels are evaluated. 
For the three intensity levels of fog and rain, as shown in Table~\ref{tab:fog_rain_levels_result}, the `Baseline' model after domain adaptation could get better detection performance compared to the `Source only' without DA, while the Proposed Method could continue to further improve the performance compared to the `Baseline' method. Alternatively, our proposed DA method could significantly mitigate the impact of fog and rain under Small, Medium, and Large intensity levels.

\begin{table*}[]
\centering
\caption{Adaptation of Different-intensity Fog and Rain: Cityscapes$\rightarrow$Foggy Cityscapes and  Cityscapes$\rightarrow$Rainy Cityscapes experiments.}
\label{tab:fog_rain_levels_result}
\resizebox{1\textwidth}{!}{%
\begin{tabular}{@{}c|c|cccccccc|c@{}}
\toprule
\textbf{Foggy} / \textbf{Rainy} & \textbf{Methodologies} & $C_{bus}$          & $C_{bicycle}$ & $C_{car}$  & $C_{mcycle}$ & $C_{person}$ & $C_{rider}$         & $C_{train}$ & $C_{truck}$ & mAP  \\ \midrule
                 & Source only      & 27.1 / 46.3          & 28.3 / 26.0    & 32.8 / 54.8 & 18.4 / 25.8   & 28.6 / 34.7   & 32.2 / 35.9          & 4.9 / 26.9  & 14.7 / 23.9  & 23.4 / 34.3 \\
                 & Baseline         & 45.4 / 49.5          & 36.7 / 32.6    & 53.5 / 58.3 & 26.0 / 31.4   & 36.1 / 36.1   & 45.9 / 41.4          & 37.1 / 43.4  & 26.3 / 35.1  & 38.4 / 41.0 \\
\multirow{-3}{*}{Large} &
  Proposed DA Method &
  \textbf{51.2} / \textbf{60.0} &
  \textbf{39.1} / \textbf{35.3} &
  \textbf{54.3} / \textbf{60.6} &
  \textbf{31.6} / \textbf{33.8} &
  \textbf{36.5} / \textbf{38.8} &
  \textbf{46.7} / \textbf{42.9} &
  \textbf{48.7} / \textbf{52.4} &
  \textbf{30.3} / \textbf{36.3} &
  {\textbf{42.3} / \textbf{45.0} } \\ \midrule
                 & Source only      & 40.6 / 47.1          & 36.9 / 26.9   & 48.9 / 54.6 & 27.3 / 23.9   & 37.9 / 34.0   & 43.2 / 38.6           & 40.4 / 30.8 & 22.4 / 30.7  & 37.2 / 35.8 \\
                 & Baseline         & 52.0 / 47.3           & 40.5 / 33.3    & 58.9 / 58.2 & 31.7 / 27.9   & 40.8 / 35.8    & \textbf{50.0} / 44.3 & 39.8 / 35.1  & 29.9 / 36.4  & 42.9 / 39.8 \\
\multirow{-3}{*}{Medium} &
  Proposed DA Method &
  \textbf{52.6} / \textbf{58.8} &
  \textbf{42.6} / \textbf{37.1} &
  \textbf{59.3} / \textbf{60.0} &
  \textbf{32.1} / \textbf{30.9} &
  \textbf{41.2} / \textbf{38.6} &
  47.5 / \textbf{45.1} &
  \textbf{48.8} / \textbf{41.0} &
  \textbf{32.5} / \textbf{39.7} &
  \textbf{44.6} / \textbf{43.9} \\ \midrule
                 & Source only      & 49.2 / 43.8          & 40.9 / 29.6   & 55.7 / 55.7 & 33.1 / 24.1   & 41.0 / 35.7   & 47.0 / 37.5         & 43.1 / 38.7  & 28.4 / 23.3  & 42.3 / 36.0 \\
                 & Baseline         & \textbf{54.1} / 42.9 & 40.6 / 34.5    & 60.3 / 59.0 & 32.5 / 30.7   & 42.0 / 36.7   & \textbf{51.0} / \textbf{43.7}  & 49.3 / 48.1  & 31.9 / \textbf{36.1} & 45.2 / 41.5 \\
\multirow{-3}{*}{Small} &
  Proposed DA Method &
  52.9 / \textbf{53.6} &
  \textbf{43.1} / \textbf{38.3} &
  \textbf{60.6} / \textbf{61.3} &
  \textbf{36.3} / \textbf{33.6} &
  \textbf{42.7} / \textbf{39.4} &
  49.4 / 42.8 &
  \textbf{54.8} / \textbf{51.4} &
  \textbf{36.5} / 35.7 &
  \textbf{47.0} / \textbf{44.5} \\ \bottomrule
\end{tabular}%
}
\end{table*}

\subsection{Adaptation of Cross Cameras} 
We conducted an experiment specifically targeting real-world cross-camera adaptation for different autonomous driving datasets with varying camera settings.
We applied our DA method for cross-camera adaptation \textit{i.e.}, Cityscapes dataset (source) $\rightarrow$  KITTI dataset (target). To accommodate the unaligned nature of the datasets, we simply removed the $L^R_{obj}$ term (Eq.~\ref{equ:reg_loss_obj}) during the adaptation process. Following the previous work~\cite{chen2018domain}, we used the KITTI training set, consisting of 7,481 images as the target domain. Specifically, we evaluated the AP of the Car category on the target domain. Table~\ref{tab:comparison_result1} demonstrated the outstanding performance of our proposed DA+ method compared to recent comparison methods.

\subsection{Feature Distribution Visualization via Adaptation} 
To investigate the capability of our proposed DA method to overcome the domain shift (clear weather $\rightarrow$ rainy/foggy weather), we visualize these domain feature distributions by utilizing t-SNE~\cite{van2008visualizing} before and after the domain adaptation in foggy and rainy weather. Fig.~\ref{fig:feature_distribution} obviously presents that our proposed DA method could align the feature distributions to bridge the domain gap (clear weather $\rightarrow$ rainy/foggy weather).

\begin{table}[htb]
\centering
\caption{Adaptation of cross cameras on Cityscapes$\rightarrow$KITTI experiment.}

\label{tab:comparison_result1}
\begin{tabular}{@{}c|c@{}}
\toprule
Methodologies         & Car AP \\ \midrule
MAF-ICCV'2019~\cite{he2019multi}     & 72.10   \\
SWDA-CVPR'2019~\cite{saito2019strong}    & 71.00 \\
ATF-ECCV'2020~\cite{he2020domain}     & 73.50   \\
ART-CVPR'2020~\cite{zheng2020cross}     & 73.60   \\
GPA-CVPR'2020~\cite{xu2020cross}     & 65.36  \\
SGA-TMM'2021~\cite{zhang2021self}      & 72.02  \\
UIT-ESwA'2022~\cite{arruda2022cross}     & 73.70   \\
ParaUDA-TITS'2022~\cite{zhang2022parauda} & 72.20   \\
IDF-TCSVT'2023~\cite{lang2022exploring}    & 74.00     \\
\midrule
Ours            &74.38  \\ 
Ours$\textbf{+}$            & \textbf{74.71}  \\ 
\bottomrule
\end{tabular}
\end{table}

\subsection{Experiments on Different Parameters}
We analyze the detection performance on different hyper-parameters in Section~\ref{Sec:Method}, \textit{.i.e}, Eq.\ref{equ:all_loss} and Eq.\ref{equ:ad_grl} for the Cityscapes$\rightarrow$Foggy Cityscapes case, and several hyper-parameters were investigated.
First of all, $mAP_{\gamma}$ can be obtained $mAP_{0.1}=42.34, mAP_{0.01}=41.30, mAP_{0.001}=41.19$, where $\gamma$  represents loss balance weight in Eq.~\ref{equ:all_loss}.
Then, in the AdvGRL (Eq.~\ref{equ:ad_grl}), the $(\alpha, \beta)$, where $\beta$  represents  the overflow threshold and $\alpha$ represents hardness threshold are set as (a) $ (0.63, 30)$, (b) $(0.63, 10)$, (c) $(0.54, 30)$, and (d) $(0.54, 10)$, where $\alpha=0.54$ is obtained by averaging the values of  Eq.~\ref{equ:img_loss} when $P_i=0.9, G_i=1$ and $P_i=0.1, G_i=0$. The corresponding detection mAP(s) are (a) 42.34, (b) 38.83, (c) 39.38, (d) 40.47, respectively.

\subsection{Visualization of Hard Examples}
\begin{figure}[t]
  \centering
  \includegraphics[width=1\linewidth]{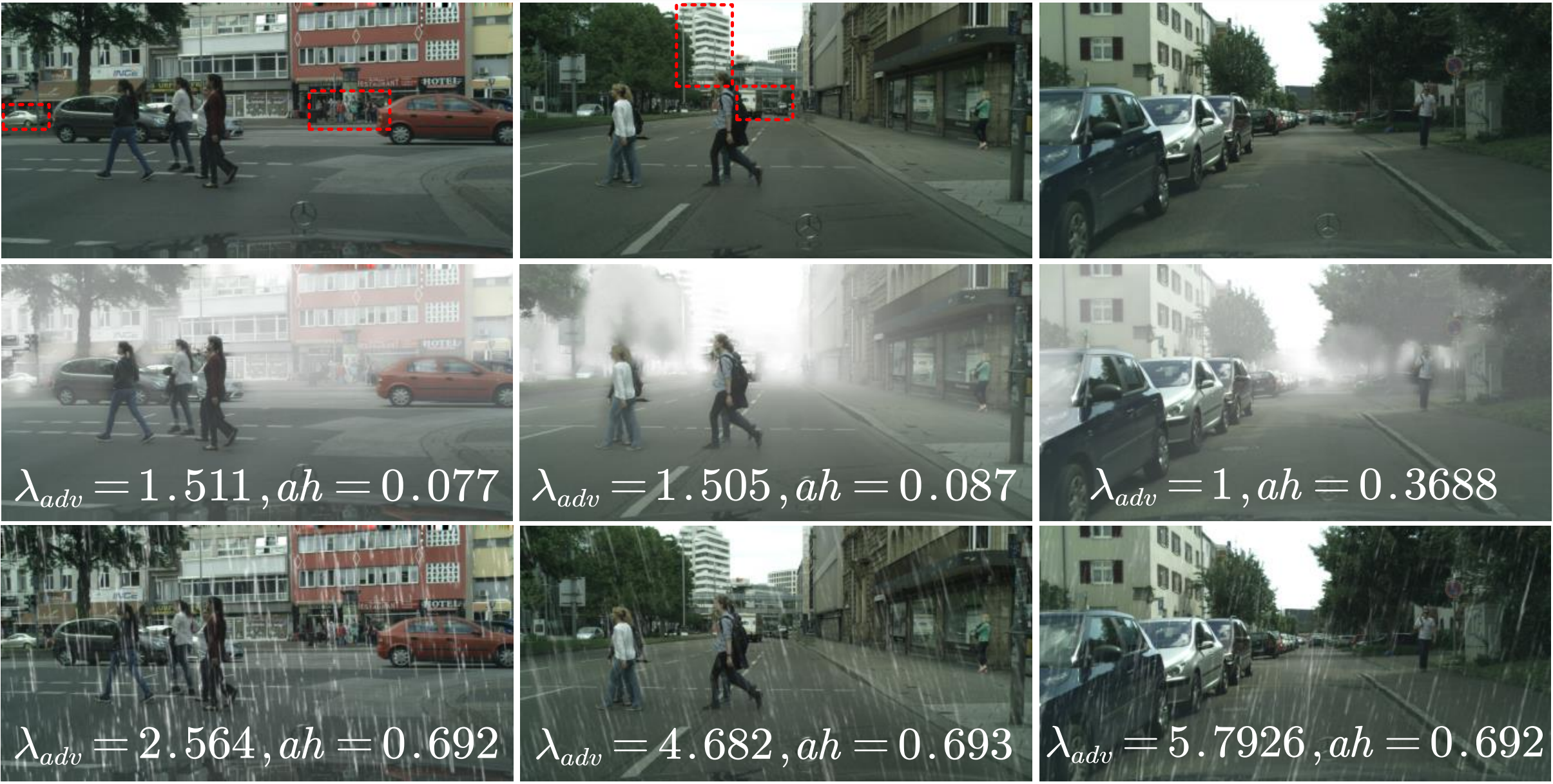}
 \caption{Visualization of hard examples mined by AdvGRL. Two mined hard examples and one easy example are shown from left to right.} 
  \label{fig:hard-examples}
\end{figure}

By utilizing  $\lambda_{adv}$ of the proposed AdvGRL, we can identify hard examples during the domain adaptation process. Fig.\ref{fig:hard-examples} illustrates some of these hard examples. We compute the $L_1$ distance between the features $F^S_i$ and $F^T_i$ obtained from the backbone of Fig.\ref{fig:da_faster_rcnn}. This distance is used as an approximation of the example's hardness ($ah$), where a smaller $ah$ indicates a harder example for transfer learning. Intuitively,
when the fog covers a larger number of objects, as illustrated by the bounding-box regions in Fig.~\ref{fig:hard-examples}, the task becomes more challenging.

\subsection{Experiments on Pre-trained Models and Domain Randomization} 

\noindent \textbf{Pre-trained Models:} In the experiment of Cityscapes$\rightarrow$Foggy Cityscapes, our proposed DA method utilizes a pre-trained Faster R-CNN as an initialization and achieves a detection mean Average Precision (mAP) of 41.3,
compared to a mAP of 42.3 achieved when our method is initialized without the pre-trained deep learning model.

\noindent \textbf{Domain Randomization:} In the Cityscapes$\rightarrow$Foggy Cityscapes experiment, we explore two approaches for domain randomization to reduce the domain shift between the source and target domains. 1) The first approach involves regular data augmentation techniques such as color change, blurring, and salt $\&$ pepper noises to construct the auxiliary domain. When our method is trained using this auxiliary domain, the detection mean Average Precision (mAP) achieved is 38.7, compared to our method's performance of 42.3 when using the auxiliary domain dataset \textit{i.e.} rain synthesis Cityscapes dataset. 2) The second approach utilizes CycleGAN~\cite{zhu2017unpaired} to facilitate the transfer of image style between the Cityscapes training set and the Foggy Cityscapes training set.
We trained a Faster R-CNN with these generated images, which got 32.8 mAP.
These findings emphasize the limitations of commonly employed domain randomization techniques in effectively addressing the DA challenge.

\section{Conclusions}\label{Sec:Conclusions}
In this paper, a novel domain adaptive object detection framework is presented, which is specifically designed for intelligent vehicle perception in foggy and rainy weather conditions. The framework incorporates both image-level and object-level adaptations to address the domain shift in global image style and local object appearance.  An adversarial GRL is introduced for adversarial mining of hard examples during domain adaptation. Additionally, a domain-level metric regularization is proposed to enforce feature metric distance between the source, target, and auxiliary domains. The proposed method is evaluated through transfer learning experiments from Cityscapes to Foggy Cityscapes, Rainy Cityscapes, and KITTI. The experimental results demonstrate the effectiveness of the proposed DA method in improving object detection performance. This research contributes significantly to enhancing intelligent vehicle perception in challenging foggy and rainy weather scenarios.




 

%




\ifCLASSOPTIONcaptionsoff
  \newpage
\fi



%


\bibliographystyle{IEEEtran}
\bibliography{Jinlong}

\end{document}